\documentclass[sigconf, 10pt]{acmart}


\usepackage{balance}
\usepackage{soul}
\usepackage{hyperref}
\usepackage{multirow}
\usepackage[utf8]{inputenc}
\usepackage[ruled, lined, longend, linesnumbered]{algorithm2e}
\usepackage{amsmath,multicol}
\usepackage{subcaption,textcomp,comment}

\usepackage{booktabs}
\usepackage{multirow}

\usepackage{paralist}
\usepackage{pifont}

\usepackage{ulem}

\usepackage{array}
\newcolumntype{L}[1]{>{\raggedright\let\newline\\\arraybackslash\hspace{0pt}}m{#1}}
\newcolumntype{C}[1]{>{\centering\let\newline\\\arraybackslash\hspace{0pt}}m{#1}}
\newcolumntype{R}[1]{>{\raggedleft\let\newline\\\arraybackslash\hspace{0pt}}m{#1}}

\newcommand{\mwx}[1]{\textbf{\color{red}{#1}}}

\newcommand{\cdq}[1]{{\color{blue}{#1}}}

\newcommand{\sys}{\texttt{FeS}\xspace}

\usepackage{wrapfig}
\usepackage{comment}
\usepackage{colortbl}

\usepackage{hyperref}

\usepackage{color,soul}
\usepackage{graphics}
\usepackage{hyperref}
\usepackage{lipsum}
\usepackage{tikz}
\usepackage{enumitem}	
\usepackage{listings} 

\usepackage{booktabs}	
\usepackage{lipsum}

\usepackage{capt-of}	

\usepackage{todonotes}  
\usepackage{subcaption} 

\definecolor{refkey}{rgb}{0,0,1}
\definecolor{labelkey}{rgb}{0,0,1}

\usepackage{cleveref}
\AtBeginDocument{\DeclareCaptionSubType{lstlisting}}
\crefname{sublstlisting}{listing}{listings}
\Crefname{sublstlisting}{Listing}{Listings}

\usepackage{mathrsfs}	
\usepackage{amsfonts}

\usepackage[export]{adjustbox} 

\usepackage{boldline}					

\usepackage{multirow}

\renewcommand{\paragraph}[1]{\vskip 3pt\noindent\textbf{#1 }}	 

  {\begin{list}{$\bullet$}%
     {\setlength{\parsep}{0pt}%
      \setlength{\topsep}{0pt}%
      \setlength{\itemsep}{2pt}}}%
  {\end{list}}
\newcommand\Noted[1]{} 

\newcommand\xzlNote[1]{\sethlcolor{yellow} \hl{xzl: #1}} 

\definecolor{darkblue}{rgb}{0.0, 0.0, 0.55}
\definecolor{mygreen}{HTML}{ADFF2F}
\definecolor{mylightgray}{gray}{0.8}

  {\begin{itemize}
	[leftmargin=0cm,
		itemindent=.3cm,
		labelwidth=\itemindent,
		labelsep=0pt,
		parsep=0.0pt,
		topsep=0.5pt,
		itemsep=0.5pt,
		align=left]
  }%
  {\end{itemize}}    

  {\begin{enumerate}
	[leftmargin=.cm,itemindent=.5cm,labelwidth=\itemindent,
		labelsep=0pt,
		parsep=1pt,
		topsep=1pt,
		itemsep=3pt,
		align=left]
  }%
  {\end{enumerate}}    

\makeatletter
\def\@copyrightspace{\relax}
\makeatother

\begin{document}

    \title[Federated Few-Shot Learning for Mobile NLP]{Federated Few-Shot Learning for Mobile NLP}
    
    \author{Dongqi Cai}
	\affiliation{
	\institution{Beiyou Shenzhen Institute}
	\country{}
	}

	\author{Shangguang Wang}
	\affiliation{
	\institution{Beiyou Shenzhen Institute}
	\country{}
	}

    \author{Yaozong Wu}
	\affiliation{
	\institution{Beiyou Shenzhen Institute}
	\country{}
	}
	\author{Felix Xiaozhu Lin}
	\affiliation{
	\institution{University of Virginia}
	\country{}
	}

	\author{Mengwei Xu}
	\affiliation{
	\institution{Beiyou Shenzhen Institute}
	\country{}
	}


\begin{abstract}     
Natural language processing (NLP) sees rich mobile applications. 
To support various language understanding tasks, 
a foundation NLP model is often fine-tuned in a federated, privacy-preserving setting (FL). 
This process currently relies on at least hundreds of thousands of labeled training samples from mobile clients; 
yet mobile users often lack willingness or knowledge to label their data.
Such an inadequacy of data labels is known as a few-shot scenario;
it becomes the key blocker for mobile NLP applications. 

For the first time, this work investigates federated NLP in the few-shot scenario (FedFSL). 
By retrofitting algorithmic advances of pseudo labeling and prompt learning, we first establish a training pipeline that delivers competitive accuracy when only 0.05\% (fewer than 100) of the training data is labeled and the remaining is unlabeled.
To instantiate the workflow, we further present a system \texttt{FeS}\xspace\footnote{
    \sys is available at https://github.com/UbiquitousLearning/FeS.
}
, addressing the high execution cost with novel designs:
(1) Curriculum pacing, which injects pseudo labels to the training workflow at a rate commensurate to the learning progress;
(2) Representational diversity, a mechanism for selecting the most learnable data, only for which pseudo labels will be generated; 
(3) Co-planning of a model's training depth and layer capacity.
Together, these designs reduce the training delay, client energy, and network traffic by up to 46.0$\times$, 41.2$\times$ and 3000.0$\times$, respectively.
Through algorithm/system co-design, \sys{} demonstrates that FL can apply to challenging settings where most training samples are unlabeled. 
\end{abstract}

    \acmYear{2023}\copyrightyear{2023}
	\setcopyright{acmlicensed}
	\acmConference[ACM MobiCom '23]{The 29th Annual International Conference on Mobile Computing and Networking}{October 2--6, 2023}{Madrid, Spain}
	\acmBooktitle{The 29th Annual International Conference on Mobile Computing and Networking (ACM MobiCom '23), October 2--6, 2023, Madrid, Spain}
	\acmPrice{15.00}
	\acmDOI{10.1145/3570361.3613277}
	\acmISBN{978-1-4503-9990-6/23/10}

    \maketitle
    

\section{Introduction}\label{sec:intro}
\begin{figure}[t]
	\centering
	 \includegraphics[width=0.49\textwidth]{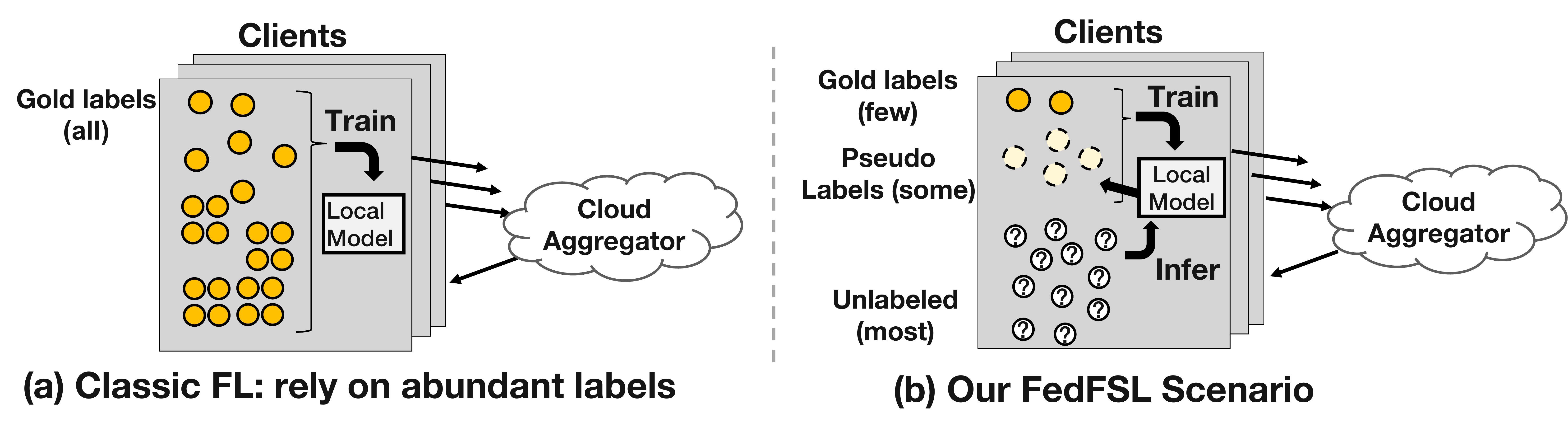}
	 \vspace{-25pt}
	\caption{Classic and few-shot federated scenarios} 
	\vspace{-20pt}
	\label{fig:bkgnd-fedfsl}
\end{figure}
\paragraph{Mobile NLP \& training data}
Mobile NLP sees rich applications on mobile devices.
Examples include auto completion, QA, and sentiment analysis~~\cite{zhang2018deep,shao2019transformer,van2019does,kim2021code, svyatkovskiy2020intellicode}. 
NLP models are trained in two phases.
(1) \textit{Pre-training} initializes a foundation model (e.g., BERT~\cite{devlin2018bert}). 
This phase learns language representations. 
(2) \textit{Fine-tuning} adjusts the model to specific NLP tasks and domains, such as to tag named identities (a task) in a user text message (a domain). 
Of NLP training, a core issue is training data. 
While pre-training is self-supervised and only needs unlabeled data, 
fine-tuning data is more difficult to obtain for the following reasons.  

First, fine-tuning often requires mobile users' private data such as their text messages. 
Fortunately, such a privacy concern is mostly addressed by federated learning (FL)~\cite{mcmahan2017communication,yang2019federated,bonawitz2019towards1, lin-etal-2022-fednlp,kairouz2021advances}, in which clients cooperate to train a model without sharing their raw data. 

A much bigger challenge is the need for data labels. 
While prior ML research tackled training with scarce labels (referred to as few-shot or zero-shot scenarios)~\cite{garcia2017few,schick2022true,chen2019closer,brown2020language,gu2021ppt}, 
the mobile environment exacerbates such scarcity, 
as most mobile users lack motivations for labeling their data (e.g., to tag which words of a text message belong to named identities)~\cite{xu2021limu} or the knowledge to do so (e.g., is a given sentence subjective or objective?)~\cite{rothe2018deep}. 
As a result, most clients are likely to have no labeled data (although they may have abundant unlabeled data); 
across all clients, only a small fraction of the total data is labeled~\cite{schick2020exploiting, van2020towards,xu2021limu}. 

\paragraph{FedFSL}
We address mobile NLP training where data labeling challenges are at their extreme. 
(1) Highly scarce data labels, e.g. less than 0.1\% of all, which are 1--2 orders of magnitude smaller than prior few-shot research~\cite{zhao2020semi, feng2022semi, diao2021semifl, jeong2020federated,fan2021federated, yu2021fedhar} (at least 4\%). 
(2) Highly skewed label availability, as opposed to most prior work that rely on 
 \textit{uniformly} distributed labels~\cite{lin-etal-2022-fednlp, lai2022fedscale,li2021hermes,shin2022fedbalancer}. 
As we will show ($\S$\ref{sec:measurement}), 
these two challenges render most prior work inadequate, resulting in significant loss in training accuracy. 
To this end, we design a runtime system for few-shot NLP learning in federated settings, 
which we refer to as FedFSL. 

\paragraph{ML building blocks}
We identify two ML algorithms as key building blocks underpinning practical FedFSL. 
(1) \textit{Pseudo labeling}~\cite{lee2013pseudo} allows to train a model $\mathscr{M}$ with a small number of labeled samples (called gold labels) together with many unlabeled samples. 
To do so, $\mathscr{M}$ makes inference on unlabeled training data, from which high-confident results (i.e., pseudo labels) are selected to further train $\mathscr{M}$.  
Such positive feedback reinforces the model's capability in decision making. 
(2) \textit{Prompt learning}~\cite{schick2020exploiting} is a recent NLP advance that better adjusts pretrained models to downstream tasks. 
With it, even a weak model is able to generate pseudo labels with less errors. 
Integrating the two algorithms, we show that a model can be fine-tuned with only tens of gold labels, while still achieving 85.8\%--97.0\% of relative accuracy as compared to a fully supervised training with thousands of labels. 

\paragraph{Challenges}
Despite of satisfactory accuracy, FedFSL can be far more expensive than standard FL, notably:

\begin{itemize}[leftmargin=*]
\item 
\textbf{Planning for FSL}. 
Our FedFSL pipeline must run inference and training in an iterative fashion. 
In an inference round, each participating client generates pseudo labels. 
Using both its gold and pseudo labels, a client runs multiple training rounds before uploading its local model updates to the cloud server. 
Having aggregated updates from many clients, the server dispatches an updated model to clients for future pseudo labeling. 

This process crucially depends on a coherent plan for inference and training: 
what pseudo labels to be injected to the learning process and at what pace. 
For instance, generating too few pseudo labels per round slows down training; 
generating too many pseudo labels, especially when the model is still weak, results in excessive erroneous labels that mislead training. 
The decision must also be dynamic, catering to different datasets and different times in a training session. 

\item 
\textbf{Excessive on-device inference}. 
After receiving an aggregated model from the server, a mobile client may run inference on \textit{all} its unlabeled data for pseudo labeling~\cite{cai2022augfedprompt}. 
However, most of the inference is in vain, as only a small fraction (the most confident) of pseudo labels will be selected for subsequent training.
The inference dominates the total energy cost, up to 87.4\% per our measurement.
The clients need an efficient mechanism for skipping generating pseudo labels for 
much of the data. 

\item 
\textbf{Training a large model on-device}. 
Language prompts rely on a large foundation model to work, 
as only a large model contains sufficiently rich knowledge amenable to extraction via prompts~\cite{liu2021pre,bommasani2021opportunities}.
Compared to models commonly used for FL, e.g., DistilBERT~\cite{sanh2019distilbert} and ALBERT~\cite{lan2019albert},
a large model such as RoBERTa-large~\cite{liu2019roberta} achieves up to 38.9\% higher accuracy, at the cost of 7.6$\times$ more computation and 4.8$\times$ more memory. 
Training with a small batch size 4 takes 75.9 seconds, 910.8 joules, and over 10 GB of memory on mainstream mobile hardware. 
\end{itemize}

\paragraph{Our solution}
We present \sys{}, the first framework that manages \underline{Fe}derated \underline{Fe}w-\underline{S}hot learning for mobile NLP.
Corresponding to the challenges above, \sys{} centers on three new designs. 

\textbf{Curriculum pacing} ($\S$\ref{sec:design-pacing}): 
To plan pseudo labeling and training, the key is to be commensurate with the learning progress. 
Intuitively, 
only as the model becomes confident via training, the client may pick increasingly more pseudo labels per round. 
Specifically, we characterize the \textit{training pace} as a configuration $\pi=\langle f,n,k \rangle$, 
where $f$ is the number of training rounds between re-generations of pseudo labels; 
$n$ is the number of participating clients; 
$k$ is the confidence threshold for selecting pseudo labels. 
Continuously weighting the model's recent learning progress against the training cost, \sys{} probes for configurations and switches to the most suitable ones, pacing through a dynamic \textit{curriculum} of training.

\textbf{On-device inference with representational diversity} ($\S$\ref{sec:design-filter}):
To reduce the inference effort, 
a client only selects its samples worth learning most and generates their pseudo labels accordingly.
How to identify such samples?
Motivated by representation learning~\cite{margatina2021active, su2022selective,reimers2019sentence}, 
our rationale is to diversify their data representations. 
For instance, the selected samples could be sentences showing varying lengths and word frequencies.
The rationale entails a lightweight implementation: 
a client estimates the \textit{approximate} representations of its samples by running inference with a low-cost proxy model; 
the client only does the estimation once, ahead of any training sessions; 
during a training session, the client selects samples for pseudo labeling by jointly considering representativeness and diversity.
We are the first to apply representational filtering in the FedFSL scenario, i.e., a training-inference collaborative pipeline.

\textbf{Co-planning of training depth and layer capacity}  ($\S$\ref{sec:design-optimization}): 
A canonical approach to efficient fine-tuning is to only train the top layers\footnote{Top layers: layers closer to a model's output~\cite{houlsby2019parameter}.} which encode task-specific knowledge while freezing the bottom layers which encode task-agnostic knowledge. 
For FedFSL however, 
we find that layer freezing alone is inadequate, resulting in inferior model accuracy. 
We attribute the observation to that 
much of the task-agnostic knowledge shall be adjusted as well in case of label scarcity. 
Therefore, \sys{} controls a model's training depth and its capacity jointly. 
Specifically, each client trains a model's top layers with full layer capacity, trains the middle layers with reduced capacity, and freezes the very bottom layers. 
The client reduces a layer's capacity by only tuning its bias while freezing its weights, a technique retrofitted from prior work~\cite{zaken2021bitfit, sun2022unfreeze,logan2021cutting}.
\sys{} is the first to explore the synergy between layer freezing and model capacity, to our best knowledge. 

It is worth noting that the above designs are compatible with existing FL optimizations such as client scheduling
~\cite{lipyramidfl, nishio2019client} 
and training data sampling~\cite{wang2021device, li2021sample}.
\sys{} can be realized as an enhancement to existing FL frameworks, enabling them to operate with scarce labels.

\paragraph{Results}
We implement \sys{} and test it on two embedded devices: NVIDIA TX2~\cite{tx2} and RaspberryPi 4B~\cite{rpi4b}. 
Large-scale training is emulated as prior FL literature does~\cite{lipyramidfl,li2021hermes,lai2020oort}.
On a diverse set of NLP benchmarks, \sys reaches $\sim$90\% relative accuracy while requiring three orders of magnitude fewer samples.
Compared to vanilla few-shot fine-tuning for NLP, 
\sys reduces the training delay from 2.1--9.1 hours to 0.1--0.4 hours (up to 46.0$\times$ reduction). 
Compared to strong baselines that use bias-only tuning~\cite{zaken2021bitfit}, \sys still reduces the delay by 4.3$\times$--14.6$\times$.
Our key designs contribute to the results significantly: 
curriculum pacing, representational filtering and depth/capacity co-planning reduce the delay by up to 3.5$\times$/3.5$\times$/62.3$\times$, respectively. 
\sys{} for the first time fine-tunes a big language model (RoBERTa-large) on mobile/embedded devices with only 8GB of RAM; it reduces the network traffic for model aggregation by 1,841.7$\times$ and per-device energy consumption by 21.7$\times$ on average.

\paragraph{Contributions}
We present the first work that investigates NLP training with scarce data labels in a federated setting. 

\begin{itemize}
\item \textbf{Algorithmic foundation}
We identify the algorithm foundation as a combination of pseudo labels and language prompts. 
Compared to training with fully labeled data,
we show it is possible to reduce the amount of labels by three orders of magnitude while still 
achieving 90\% of the relative accuracy. 

\item \textbf{System designs}
We tackle the high cost of FedFSL with novel designs: 
representational diversity (which optimizes inference), co-planning of learning depth and capacity (which optimizes training), and curriculum pacing (which orchestrates the two). 

\item \textbf{Experimental evaluation}
Through experiments on real hardware, for the first time we show it is both desirable and practical for mobile devices to train NLP models -- even with scarce labels. 
\end{itemize}

    \section{Motivations} \label{sec:bkgnd}


\subsection{Mobile NLP and Its Obstacles} \label{sec:bkgnd-fednlp}

This work is concerned with NLP model \textit{fine-tuning}: adapting a foundation model to various downstream tasks such as text classification and sequence tagging. 
While the foundation model is pre-trained by big companies \textit{once}, the subsequent fine-tuning recurs for individual tasks and involves mobile clients. 
Therefore, fine-tuning has a strong impact on both the model accuracy and client efficiency. 

\textbf{Explored: fine-tuning is often privacy-sensitive.}
It relies on domain samples that are often generated by users, such as user reviews, messages, or emails.
Collecting them to the cloud for training raises privacy concerns and is heavily regulated~\cite{voigt2017eu,pardau2018california}.
In response, federated learning~\cite{mcmahan2017communication,yang2019federated} is the de facto approach that trains models with good accuracy without data sharing. 
Training NLP models in a federated setting is referred to as FedNLP~\cite{lin-etal-2022-fednlp,cai2022autofednlp}. 

\textbf{Unexplored: fine-tuning is often few-shot.}
While training samples on clients can be abundant, the \textit{labeled} ones are often scarce. 
To exacerbate the problem, the numbers of labels could vary drastically across clients. 
Such skewed label distribution, combined with the non-IID data distribution nature in FL (e.g., skewed class distribution~\cite{li2022federated,lin-etal-2022-fednlp}), could further degrade the fine-tuning accuracy.
The causes for label scarcity are fundamental. 

\begin{itemize}[leftmargin=*]
    \item \textit{Users lack willingness.} Each sample is accessible to only one client user who can label it.
    Reports show that most users are reluctant to label their data~\cite{yang2018improving,fan2021federated}.
    This is fundamentally different from traditional centralized or crowd-sourced data labeling services that can recruit highly specialized data labelers~\cite{li2017crowdsourced}.
    
    \item \textit{Users lack expertise.} Data labeling for certain NLP tasks require domian-specific knowledge, which most users do not possess, e.g., cross-lingual transfer~\cite{ponti2020xcopa}, Q\&A~\cite{yang2019end}, or biomedical text corpora understanding~\cite{zhu2018gram}.
    
    \item \textit{Diverse NLP tasks.} Downstream tasks are emerging over time, e.g., new domains, topics, or data distributions. Asking users to label a large amount of data for each task is tedious, inefficient, and impractical.

    \item \textit{Mislabels are not uncommon.} Mislabels are common in real world, e.g., 6\% in the well-established ImageNet or even more than 10\% in other crowd-sourced datasets~\cite{northcutt2020pervasive}.
    In FedFSL, since the labels from end users are merely impossible to be verified, we expect an even higher ratio of mislabels, which can significantly harm the model quality.
    Instead, trainers might only use the labels from very few, highly trustworthy people.
    
\end{itemize}

In essence, we argue that few-shot is a more realistic way to depict NLP training, a scenario we call as FedFSL. 
Unfortunately, FedFSL, in particular its system implications, are rarely investigated -- 
in comparison, prior FL literature assumes abundant data labels (at least hundreds of thousands) uniformly distributed across clients. 

\subsection{Key Algorithm Blocks for FedFSL}\label{sec:algo-blocks}

In ML literature, there are two complementary approaches to address the few-shot issue.
One is to exploit the abundant \textit{unlabeled} data across clients.
The other is to boost the model's ability of learning from few samples.
For each approach, we identify a technique; together, they form the algorithmic foundation for FedFSL. 

$\bullet$ \textbf{Pseudo labeling~\cite{lee2013pseudo} allows training with few/zero labeled samples and many unlabeled samples}. 
As it trains a model, 
the trainer makes the current model infer on unlabeled data, and uses the inference results (i.e. pseudo labels) as if they are true labels for successive training. 
The efficacy of pseudo labeling has been established both empirically \cite{zhang2021flexmatch, cascante2021curriculum} and theoretically \cite{lee2013pseudo, arazo2020pseudo}. 
Intuitively, it works because training with pseudo labels encourages the model to learn a decision boundary that lies in a region where the example density is lower. 
Often, such a decision boundary yields good generalization performance (i.e. higher model accuracy), even though the true labels of individual samples remain unknown. 
In ML's lingo, such a training strategy roots in \textit{entropy regularization}: the resultant model will make a prediction on unlabeled data with low class overlap (e.g. ``great'':0.9, ``bad'':0.1 rather than ``great'':0.6, ``bad'':0.4) and therefore low entropy.  
Pseudo labeling also tackles the challenge of skewed distribution of label classes.
Since pseudo labeling involves more clients and therefore more diversified label classes for training, the fine-tuned model is likely to be more unbiased and accurate.



$\bullet$ \textbf{Prompt learning}~\cite{liu2021pre} is a powerful NLP technique that boosts accuracy in model fine-tuning, which is 
commonly used in few-shot scenarios~\cite{schick2020exploiting,liu2021p,gu2021ppt,logan2021cutting}.
For FedFSL, we find prompts crucial to the early stage of a training sessions, when the model is weak and can barely generate useful pseudo labels. 

Given a task, standard NLP fine-tuning (without prompts) trains a new classification layer from scratch, which requires supervision from substantial labeled data. 
For example, suppose we fine-tune a foundation model to classify YELP reviews~\cite{zhang2015character} and are only given two labeled samples~\cite{schick2020exploiting}: 

\begin{itemize}[leftmargin=*]
    \item[$\blacksquare$] \textit{T$_{1}$} (label=L1): ``Most delicious pizza I've ever had.''
    \item[$\blacksquare$] \textit{T$_{2}$} (label=L2): ``You can get better sushi for half the price.''
\end{itemize}

With such few samples, training a usable classification layer is impossible. 
Consider an unlabeled example: 

\begin{itemize}[leftmargin=*]
    \item[$\blacksquare$] \textit{T$_{3}$} (label=?): Pizza was good. Not worth the price.
\end{itemize}

The model may predict T3's class probabilities closer to L1 if the task is to classify user satisfaction, or closer to L2 if the task is about whether price is mentioned. 
But without such a task description, the model can only randomly guess and generate an error-prone pseudo label. 

To fix the problem, the insight of prompt learning is that the foundation model already encodes knowledge for performing various tasks; it just needs a prompt that describes what the task is. 

In the example above, if ML developers augment \textit{all} samples with a leading prompt, e.g. T3 becomes:

\begin{quote}
    \small
    "\textbf{It was <MASK>}. Pizza was good..."
\end{quote}

Then all input samples are reformulated as cloze-style phrases, 
which are exactly what the foundation model was pre-trained for -- to predict missing words (masked out) in text. 
Next, the predicted masked word is mapped to a class, yielding a label. 
For instance, a predicted word ``terrible'' will map to label 1 and ``great'' will map to label 5. 
Compared to initializing a whole classification layer from scratch, 
the foundation model with prompts requires less finetuning before it can output labels with higher accuracy; 
in FedFSL, this means that the pseudo labels are less erroneous.

More formally, a cloze question is called a pattern and the mapping from words to classes is done by a verbalizer. 
Given a task, there exist multiple possible pattern/verbalizer pairs. 
The training loss is the cross-entropy between the correct answer and the distribution of probabilities among the tokens in the verbalizer. 
See recent surveys on prompt learning~\cite{liu2021pre, bommasani2021opportunities} for more details.

\begin{figure}[t]
	\centering
	 \includegraphics[width=0.48\textwidth]{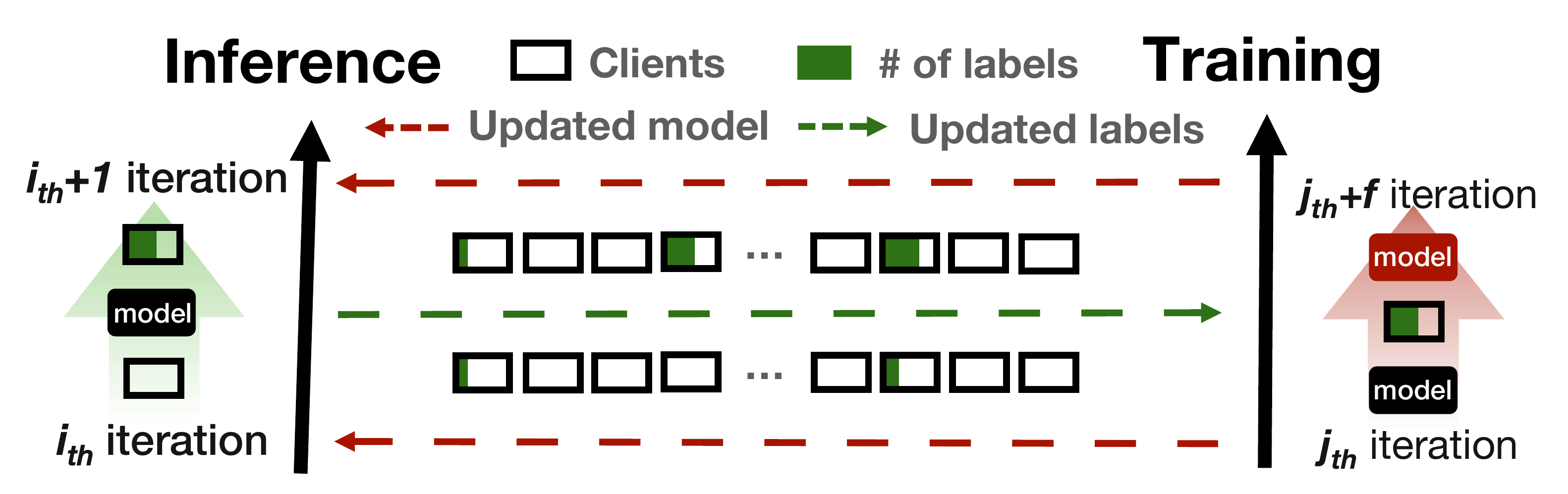}
	 \vspace{-20pt}
	\caption{Workflow of FedFSL with pseudo labeling.} 
	\vspace{-20pt}
	\label{fig:bkgnd-fedfsl}
\end{figure}

\subsection{Our System Model}

\textbf{System model: FedFSL workflow enhanced with above techniques}
We notice that pseudo labeling and prompt learning can well orchestrate and be complimentary to each other:
pseudo labeling heavily relies on the initial model accuracy to get enough, correct labels, for which prompt learning can help;
in turn, prompt learning's ability is limited to the few number of data labels and especially their skewed distribution, for which pseudo labeling can help.
Therefore, we construct an enhanced FedFSL workflow by orchestrating the two techniques atop FedNLP, shown in Fig.~\ref{fig:bkgnd-fedfsl}.
This enhanced workflow is the algorithmic foundation of our future design, and is still dubbed as FedFSL for simplicity.

Our goal is to fine-tune a pre-trained language model $\mathscr{M}$ based on distributed clients' data.
We assume that each client has a tiny training set with labels $\mathcal{T}$ (typically < 10) and a much larger set of unlabeled samples $\mathcal{D}$ (typically > 1000). 
In general, \sys consists of two loosely-coupled runtimes residing in a central server.

\begin{itemize}[leftmargin=*]
\item \textbf{Inference runtime} that continously generates new pseudo labels on clients.
Per $f$ training rounds, it dispatches the global $\mathscr{M}$ to $n$ clients, where the model exhaustively inferences on each local unlabeled data $\hat{x} \in D$ and generates a pseudo label $\hat{y}$.
The data with the top $k$ highest confidence (i.e., $logits$) are added as training samples.
In subsequent training rounds, pseudo labels are treated equally as the gold labels.
The pseudo labels that are generated in previous rounds will also be re-labeled to avoid forgetting events~\cite{toneva2018empirical}.
The above hyper-parameters $<f,n,k>$ indicate how inference runtime paces.
\item \textbf{Training runtime} that follows a typical federated learning workflow to fine-tune $\mathscr{M}$.
Per round, the runtime dispatches the global $\mathscr{M}$ to a random set of clients with at least one gold or pseudo label.
The on-device training is assisted with prompts, provided by the trainers either in hand-crafted or automatic manner~\cite{gao2020making, gu2021ppt, liu2021pre,li2021prefix, liu2021p}.
The updated models are then aggregated (default FedAvg protocol~\cite{mcmahan2017communication}) on the server as the new global $\mathscr{M}$.
The process continues till $\mathscr{M}$ reaches a satisfactory accuracy.
Notably, such a design is compatible with prior FL literature on client/data sampling~\cite{li2021hermes,lipyramidfl, xu2020client,lai2020oort, li2021sample}, privacy enhancements~\cite{dwork2008differential, zhang2020batchcrypt}, and communication optimization~\cite{wu2018error, abdelmoniem2021towards, wangni2018gradient}.
\end{itemize}

\subsection{Experimental Observations}\label{sec:measurement}

Based on the FedFSL workflow presented above, we perform a set of early experiments on its performance.
The results highlight the two sides of a coin: a satisfactory model accuracy yet huge resource cost on clients.


\begin{table}[t]
    \centering
    \scriptsize
    \begin{tabular}{l|ccccc}
    \toprule
    \textbf{Dataset}       &  \begin{tabular}[c]{@{}c@{}}\textbf{Full-set}\\ \textbf{(oracle)}\end{tabular} &
    \begin{tabular}[c]{@{}c@{}}\textbf{Vanilla-}\\ \textbf{FedFSL}\end{tabular} &
    \begin{tabular}[c]{@{}c@{}}\textbf{Prompt-}\\ \textbf{Only}\end{tabular} &
    \begin{tabular}[c]{@{}c@{}}\textbf{Pseudo-}\\ \textbf{Only}\end{tabular} &
    \begin{tabular}[c]{@{}c@{}}\textbf{Both}\\ \textbf{(Ours)}\end{tabular} \\ 
    \midrule
    \texttt{AGNEWS} (skewed) & 93.0     & 64.8$\pm$3.1           & 68.4$\pm$2.4           & 67.5$\pm$1.3        & \textbf{90.2}$\pm$0.5    \\
    \texttt{MNLI} (skewed)   & 85.0     & 37.7$\pm$5.6          & 42.4$\pm$5.8           & 42.7$\pm$6.3           & \textbf{77.4}$\pm$1.2    \\
    \texttt{YAHOO} (skewed)  & 78.0     & 24.4$\pm$10.3           & 41.8$\pm$4.3           & 31.0$\pm$2.0           & \textbf{66.9}$\pm$1.1    \\
    \texttt{YELP-F} (skewed)   & 70.0     & 38.3$\pm$8.8           & 51.2$\pm$1.8           & 45.7$\pm$4.4           & \textbf{58.2}$\pm$2.4    \\\midrule
    \texttt{YELP-F} (uniform)  & 70.0        & 54.0$\pm$0.1           & 58.1$\pm$1.5         & 57.0$\pm$2.2           & \textbf{61.9}$\pm$0.7  \\
    \bottomrule
    \end{tabular}
    \caption{Convergence accuracy with 64 gold labels.
    ``Full-Set'' assumes every data is labeled (an oracle case).
    ``skewed'' means the gold labels are located on few clients instead of uniformly distributed across clients.}
    \vspace{-20pt}
    \label{tab:eval-softlabel}
\end{table}


    
\textit{Observation-1: FedFSL achieves satisfactory accuracy with scarce data labels; for which both pseudo labeling and prompt learning are indispensable.} 
Table~\ref{tab:eval-softlabel} shows the convergence accuracy of RoBERTa-large~\cite{liu2019roberta} on 4 popular NLP datasets\footnote{You can find a detailed description of the datasets in $\S$\ref{sec:eval-setup}.}.
With only 64 data labels (0.005\%--0.05\% of the total dataset), FedFSL achieves 85.8\%--97.0\% relatively convergence accuracy to the full-set fine-tuning that assumes all data samples are labeled.
The accuracy could be further boosted by involving more data labels.
Neither pseudo labeling nor prompt learning alone is enough to exhibit a usable accuracy.
With only one of them, the relative convergence accuracy is 40\%--74\%.
Furthermore, the skewed label distribution challenges the task:
on \texttt{YELP-F}, a vanilla FedNLP method results in much higher accuracy when the labels are uniformaly distributed;
neverthelss, the challenge is mostly addressed by FedFSL that achieves satisfactory accuracy in both cases.

\begin{figure}[t]
	\centering
        \includegraphics[width=0.5\textwidth]{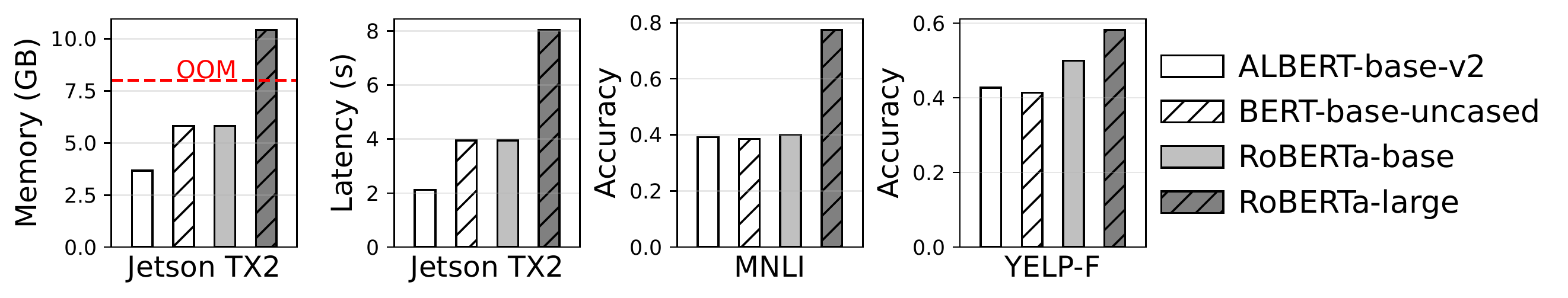}

	 \vspace{-10pt}
	\caption{
		FedFSL convergence performance with different models and datasets. Batch size: 4.
		} 
	\vspace{-15pt}
	\label{fig:eval-model}
\end{figure}
\textit{Observation-2: FedFSL incurs huge system cost.}
Our experiments highlight the excessive system cost (Figure~\ref{fig:eval-model}), as against the commonsense that few-shot learning is usually fast and lightweight~\cite{fan2021federated, zhang2020revisiting, guha2019one, chen2019closer}.
For example, training RoBERTa-large on \texttt{AGNEWS} takes 3.3 hours to converge, 7.3 million Joules of energy, 68.4 GBs of network transmission, and 10.4 GB peak memory.
The cost is about 1.4$\times$ higher than a full-set supervised FedNLP process on the same model and dataset.
We then dive deeper into the implications behind and identify three challenges for a resource-efficient FedFSL system.

$\bullet$ \textbf{Orchestrating training and inference}
FedFSL has two coupled components: a federated learning runtime that continuously updates a global model; an inference runtime that keeps generating pseudo labels.
The two components must be paced harmoniously:
the inference runtime generating too few pseudo labels could slow down the training;
otherwise, generating too many pseudo labels could lead to resource waste or even excessive erroneous labels, especially when the global model is still weak.
A mechanism to orchestrate the two components must be dynamic to fit the model learning progress.

$\bullet$  \textbf{Prompt learning needs large NLP model.}
Compact NLP models have been proposed for mobile scenarios with acceptable accuracy degradation.
However, our experiments demonstrate that a large, fully-fledged foundational language model is demanded in FedFSL.
As shown in Figure~\ref{fig:eval-model}, on \texttt{MNLI} and \texttt{YELP-F}, a large RoBERTa-large model (24 transformer blocks) can reach 91\% and 88\% relative accuracy to full-set fine-tuning, while Bert-base (12 transformer blocks) can only obtain 45\% and 59\%.
The rationale is that a large language model also encodes rich knowledge for downstream tasks, from which hand-crafted prompts can better extract useful information within limited data labels.

$\bullet$ \textbf{Excessive on-device inference to obtain trustful pseudo labels.}
Pseudo labeling typically requires performing inference on all unlabeled data, from which the most confident/valuable ones can be selected for future training.
Note that the inference is not one-pass, as the model is continuously updated each round.
According to our experiments, up to 87.4\% of the total energy cost on clients is attributed to the inference.
As will be shown in $\S$\ref{sec:design-filter}, randomly filtering out a large portion of samples to speed up pseudo labeling will degrade accuracy significantly.

    \section{\sys Design}\label{sec:design}

Atop the FedFSL workflow presented in $\S$\ref{sec:algo-blocks}, \sys introduces three key techniques to make it systematically practical by addressing each of the challenges raised in $\S$\ref{sec:measurement}.



\subsection{Curriculum Pacing} \label{sec:design-pacing}

\sys proposes to progressively speed up the pseudo labeling speed, i.e., adding more pseudo labels at a higher frequency.
The rationales are two folds.
First, at the beginning of FedFSL, the language model is relatively weak and the produced labels are prone to be erroneous, from which we better pick only the very confident ones;
as the training progresses, the model becomes accurate enough to produce trustful pseudo labels in faster speed.
Second, as the model gets more accurate, the utility of each data sample to further enhance the model diminishes.
It demands recruiting more data to sustain an effective learning progress.

\begin{figure}[t]
    \centering
    \begin{minipage}[b]{1\textwidth}
        \begin{minipage}[b]{0.25\textwidth}
            \includegraphics[width=1\textwidth]{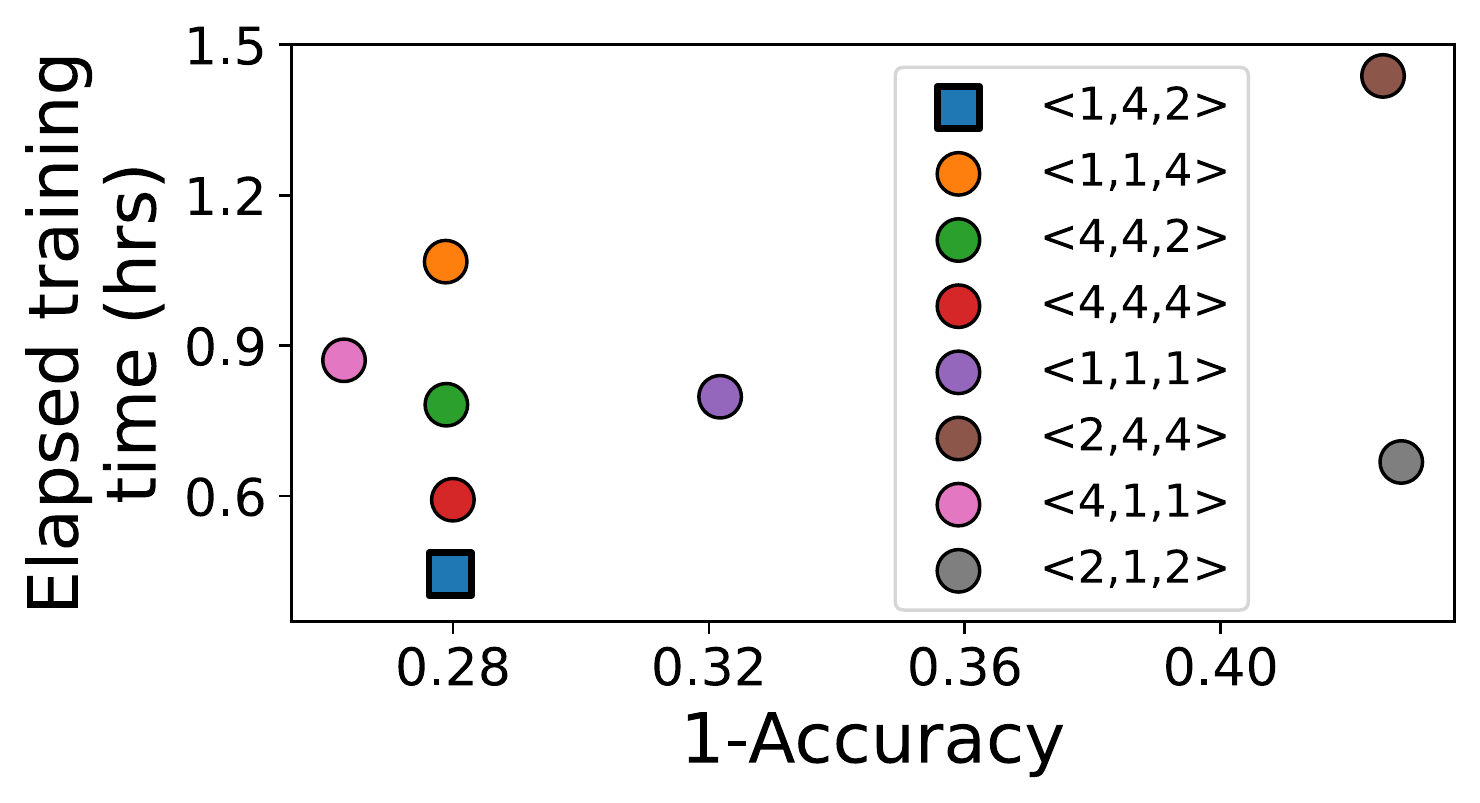}\vspace*{-1pt}
            \vspace{-15pt}
            \subcaption{Impacts of configuration}\label{fig:design-configuration-impact}
        \end{minipage}
        ~
        \begin{minipage}[b]{0.23\textwidth}
            \includegraphics[width=1\textwidth]{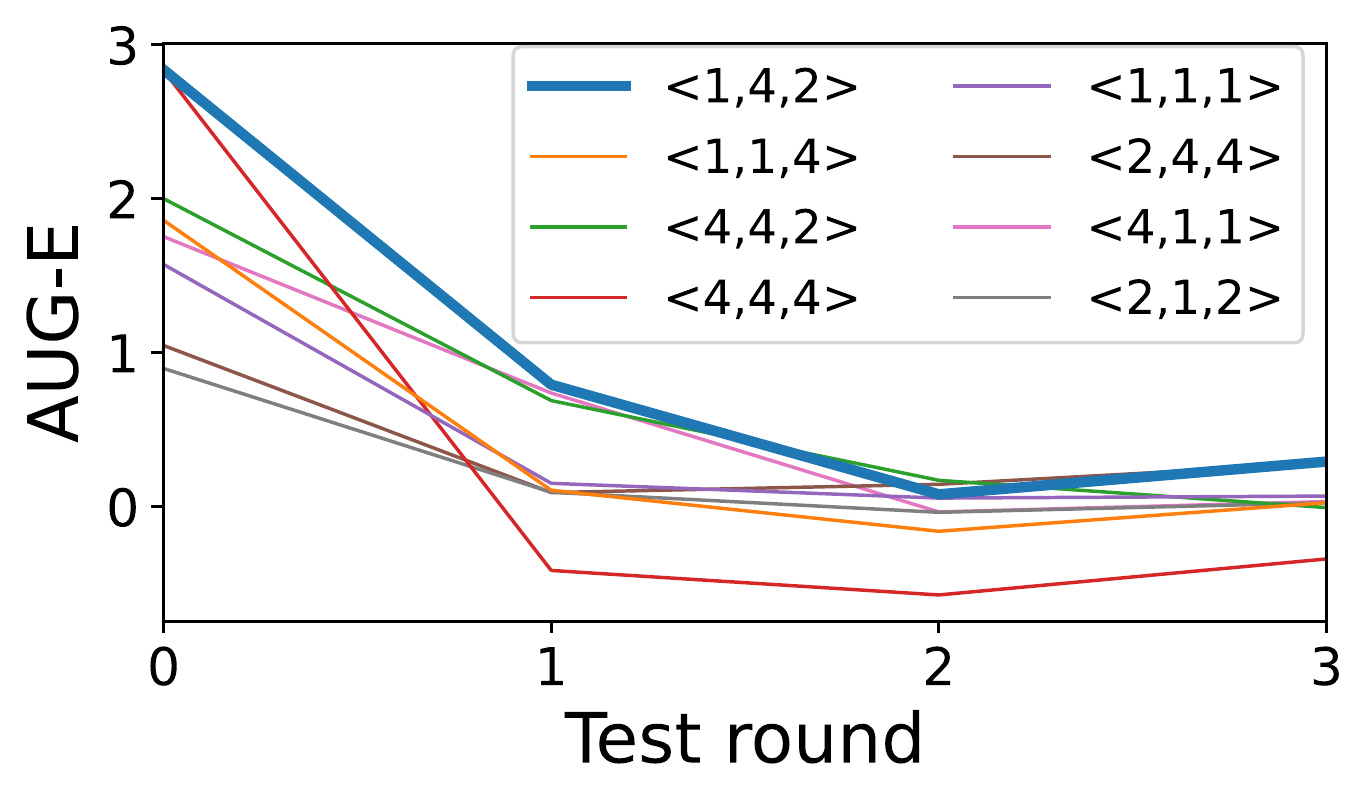}
            \vspace{-15pt}
            \subcaption{AUG-E across rounds}
        \end{minipage}
    \end{minipage}
    \vspace{-20pt}
    \caption{AUG-E metric at early rounds (a) can help identify the pacing configurations that perform well in end-to-end training (b) regarding both accuracy and system cost (the higher AUG-E, the better).
    $<f,n,k>$ is the pacing configuration.
    Dataset: MNLI~\cite{williams2017broad}.
    }
    \label{fig:design-configuration}
    \vspace{-20pt}
\end{figure}

\paragraph{The configuration space.}
\sys further probes into more detailed pacing configurations, and distills three key parameters:
$f$ indicates frequency of updating pseudo labels, i.e., the number of training rounds before the next pseudo labeling;
$n$ indicates the number of clients selected to perform pseudo labeling;
$k$ indicates the curriculum ratio (linear increase) of data selected as pseudo labels for the subsequent training per selected client.
As long as $k$ is positive (e.g., 1\%), the pseudo labeling speeds up (e.g., 1\% selected at 1st round, 2\% selected at 2nd round, etc).
For example, <f,n,k>=<5,2,1> means inference line would provide 1\% more pseudo labels compared to the prior updating from 2 clients every 5 rounds.
The three parameters form a configuration (denoted as $<f,n,k>$) that can flexibly control the relative pacing between the training and inference.
As shown in Figure~\ref{fig:design-configuration-impact}, various configuration leads to huge accuracy and cost tradeoff.
Meanwhile, the best configuration varies across datasets and models, i.e., no silver-bullet configuration.

\paragraph{Lightweight configuration searching}
To search for an effective configuration with low cost, we define a new metric \textit{augment efficiency} (AUG-E) to measure the gradient of the time-to-accuracy curve.:
\begin{align}
    AUG \text{-} E(f,n,k) = \frac{\eta  \Delta(acc)}{C_{infer}(f,n) + \theta \cdot C_{train}(k)} 
\end{align}
where $C_{infer} = l_{i} \cdot n/{f}$ and $C_{train} = l_{t} \cdot k$. 
Here, $l_{i}$ stands for inference latency and $l_{t}$ stands for training latency per batch.
AUG-E takes both (accuracy) gain and the cost for this gain into account.
Higher the AUG-E, more accuracy benefit is brought from pseudo labeling cost.

\begin{algorithm}[t]
	\footnotesize
	\SetAlgoNoEnd
  \SetKwProg{Fn}{Function}{~}{end}
  \SetKwData{Left}{left}\SetKwData{This}{this}\SetKwData{Up}{up}
  \SetKwFunction{MIN}{MIN}\SetKwFunction{MAX}{MAX}\SetKwFunction{LENGTH}{LENGTH}
  \SetKwInOut{Input}{input}\SetKwInOut{Output}{output}\SetKwInOut{Variable}{Variable}

  \Input{
      Target accuracy, $acc$;\\
      Initial configuration list, $l$;\\
      Number of candidate configurations, $t$;\\
      AUG-E threshold, $s$;\\
      Trial round, $r$.
    }

  \Output{
      Fine-tuned model, $\Theta_{i}$~(i=1,2\dots).
    }
  \BlankLine

  \Fn{Cloud\_controller():}{
    Iteration $i$=0;\\
    $l_{win}$, $l_{candidate}$, $i$ $\leftarrow$ Switch($i,l$); // initial pace configuration\\
      \While{Eval($\Theta_{i}$) $<$ $acc$}
      { 
        Pacing($i$, $l_{win}$); // training and labeling concurrently \\
        $E$ $\leftarrow$ Compute AUG-E; // accuracy degradation detect\\
        
        \If{$E$<$s$}{
          $l_{win}$, $l_{candidate}$, $i$ $\leftarrow$ Switch($i,l_{candidate}$);\\
        }
        }
        Exit training.
      }

    \Fn{Switch($i$, $list$):}{
      $i_{tmp}$ $\leftarrow$ $i$;\\
            \For{$l$ in $list$}{
              \While{$i<i_{tmp} + r$}{
                Pacing(i, $l$); i++;\\
              }
              $E$ $\leftarrow$ Compute AUG-E;\\
          }
          $l_{win}$, $l_{candidate}$ $\leftarrow$ Configurations with highest $E$ and top-$t$ $E$.
    }

  \Fn{Pacing($i$, $l$):}{
    $G_{train}$, $G_{label}$ $\leftarrow$ Selects clients groups separately;\\
      Send model $\Theta_{i}$ and configuration $l$ to $G_{train}$ and $G_{label}$;\\
      \textbf{Parallel}: $Client\_labeling(i)$, $Client\_training(i)$;\\

      $\Theta_{i+1}$ $\leftarrow$ Aggregate receive updated model from $C_{train}$.
  }

  \Fn{Client\_training(i):}{
      $\Theta_{i+1}(n)$ $\leftarrow$ Update received model $\Theta_{i}$ on local labeled data; \\
      Send updated model $\Theta_{i+1}(n)$ to cloud.
  }

  \Fn{Client\_labeling(i):}{
    $labels$ $\leftarrow$ Generate pseudo labels per pacing configuration.}

  \caption{Our Pacing Configurator}\label{alg:pacing}
  
\end{algorithm}

Algorithm~\ref*{alg:pacing} describes how \sys leverages AUG-E for configuration searching in details.
At the beginning, \sys tries each configuration from an initial list we hand-picked through extensive offline experiments (default size 32).
After only a few rounds, \sys evaluates those configurations using AUG-E:
the best one will be selected for future pseduo labeling runtime (line 3, 10--16);
the top-$t$ (default 8) ones are packed into a candidate list for future update in case (see below).

To understand how AUG-E helps predict those efficient configurations ahead, we profile 8 random configurations $<f,n,k>$ and show their convergence performance and AUG-E metric on \texttt{MNLI} dataset~\cite{williams2017broad}.
Figure~\ref{fig:design-configuration} shows that the most effective configurations with higher accuracy and smaller training time would obtain a relatively higher AUG-E score in the early stage of searching.
For example, the configuration <1,4,2> with the highest AUG-E converges at 72.0\% accuracy (2nd highest) within 0.4 hours (fastest).
Therefore, we can use AUG-E as an indicator to the end-to-end performance of different pacing configurations.

\paragraph{Configuration switching}
In practice, we observe few cases that the picked configuration performs badly as training goes on.
To mitigate the impacts of those corner cases, inspired by \cite{cai2022autofednlp}, \sys adopts a configuration switching mechanism at online.
As described in Algorithm~\ref*{alg:pacing},
once training alarms due to sharp accuracy degradation (line 6--8), i.e., AUG-E is below zero or extremely low, \sys seeks to switch the pacing configuration (usually speeds up the inference).
More specifically, \sys repeats the configuration searching as it does at the beginning rounds (line 10--16), but only with the short top-$t$ list of candidates that are proven to be relative effective as discussed above.

\paragraph{Cost analysis}
The cost of configuration searching is negligible as it spans only a few beginning rounds (typically 5);
and online switching on top-$t$ configurations rarely happens (typically < 2).
Please note that the trials on different configurations could be amortized by leveraging the large amount of idle devices in federated learning~\cite{yang2021characterizing,bonawitz2019towards1}.

\paragraph{Wrongly labeled data}
If malicious clients are unluckily selected, \sys could encounter unexpected behavior such as low AUG-E score.
Those wrongly labeled data will be flagged as `unlabeled'.
Subsequently, these data points undergo re-labeling using our pseduo-labeling mechanism.
Another advantage of \sys is that it requires only a small number of labeled training data, often in the tens.
Consequently, it becomes easier for the cloud to identify trustworthy clients whose data labels are more likely to be accurate.

\subsection{Representational Filtering} \label{sec:design-filter}

To circumvent the exhaustive labeling (model inference) of all local data, an intuitive approach is to early filter the data that is likely to contribute minimally to subsequent training.
There are two key questions to be answered:
(1) what metrics shall be used to quantify the value of a sample if it is important for training;
(2) How can these metrics be efficiently extracted for each sample? Ideally, this process should be decoupled from the NLP model that is being continuously trained, allowing for an offline, one-pass operation.

\paragraph{Representativeness- and diversity-aware score}
The key idea of \sys is to jointly consider two data aspects: \textit{representativeness} helps many text instances to find similar demonstrations, thus reducing duplicate labeling (inferring) cost for similar samples;
\textit{diversity} guarantees enough statistical utility, thus increasing the total convergence performance.

To do so, \sys first computes a vector representation for each unlabeled training instance (a sentence) $x \in \mathcal{X}$, by averaging the output vector for each of its word using the proxy model discussed below. 
For each sample $x$, we sort its $k$ most similar samples in terms of the cosine similarity between the embedding vectors.
Those example ids are denoted as $\mathcal{V}(x)$.  
We denoted representative vector as $\mathcal{E}(u) = \{x \mid u \in \mathcal{V}(x), x \in \mathcal{X}\}$.
$|\mathcal{E}(u)|$ means the quantity of samples that sample $u$ is similar to.
Bigger the $|\mathcal{E}(u)|$, larger the representativeness.

\begin{figure}[t]
    \centering
    \begin{minipage}[b]{1\textwidth}
        \begin{minipage}[b]{0.24\textwidth}
            \includegraphics[width=1\textwidth]{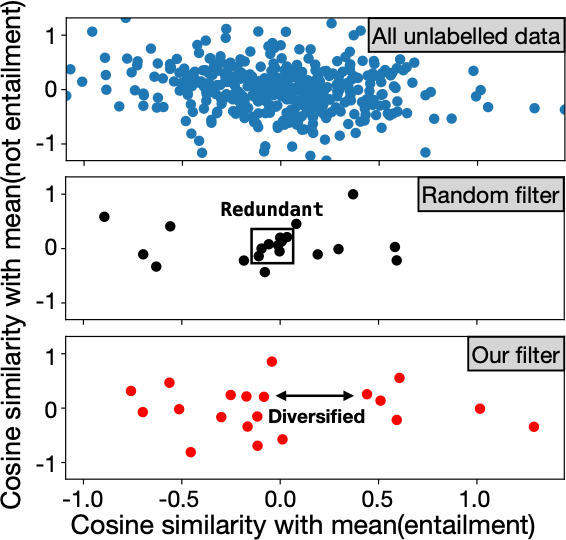}
            \subcaption{Representative diversity} \label{fig:design-filter-diversity}
        \end{minipage}
        ~
        \hspace*{2pt}
        \begin{minipage}[b]{0.24\textwidth}
            \includegraphics[width=0.82\textwidth]{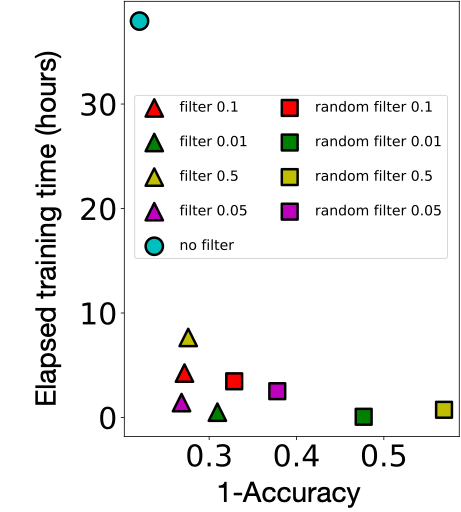} 
            \subcaption{End-to-end performance} \label{fig:design-filter-performance}
        \end{minipage}
    \end{minipage}
    \vspace{-20pt}
    \caption{
    \sys's representational filter helps find out a small portion of samples (20\%) that are diversified in text semantic as compared to random filtering (a).
    In end-to-end experiments (b), it accelerates convergence by reducing pseudo labeling cost with negligible accuracy loss.
    Dataset: \texttt{YAHOO}~\cite{zhang2015character}.
    }
    \label{fig:design-filter}
    \vspace{-20pt}
\end{figure}

Now let $\mathcal{L}$ and $\mathcal{U}$ denote the sets of already chosen samples and remaining samples, respectively. 
Initially, $\mathcal{L}=\emptyset$. 
Every remaining sample $u \in \mathcal{U}$ is scored as below:
\begin{align}
    \operatorname{score}(u)=\sum_{x \in \mathcal{E}(u)} \rho^{-|\{\mathcal{V}({x}) \cap \mathcal{L} \}|}, \quad \rho>1
\end{align}
where $\rho$ discounts $x$ that is close to the already selected instances, thereby encouraging diversity. 
A sample with higher $\operatorname{score}(u)$ would be preferred for labeling.
Once it is selected, it will move from $\mathcal{U}$ to $\mathcal{L}$.
$\operatorname{Scores}$ would be updated subsequently.
According to Figure~\ref{fig:design-filter-performance}, up to 95\% online inferring cost could be saved without harming convergence performance apparently.
This speeds up the end to end performance significantly.

\paragraph{Low-cost proxy model}
Existing importance-based filtering methods observe training loss~\cite{shin2022fedbalancer}, or weight norm~\cite{alain2015variance, katharopoulos2018not} to measure the importance of each training data.
Those methods are not feasible in FedFSL that lacks data labels.
Inspired by sentence annotating task~\cite{su2022selective} and sentence-pair regression task~\cite{reimers2019sentence}, we propose to use a BERT-like model to obtain a feature representation for each data sample.

We use Sentence-BERT~\cite{reimers2019sentence} as the proxy model, which is priorly used for sentence similarity comparison.
It is a medium-sized NLP model that consists of 12 transformer blocks, which is notably smaller than the NLP model being trained in FedFSL as discussed in $\S$\ref*{sec:algo-blocks}.
Furthermore, the proxy model is trained offline and independent from the NLP model to be trained through FedFSL.
Therefore, computing the vector representation of each data is one pass and incurs much less cost than performing inference each round. 
For example, when computing the vector representation of YELP-F~\cite{zhang2015character}, it incurs only an additional 8.4\% time cost compared to the FedFSL training process. Moreover, this cost can be further amortized across multiple FedFSL tasks since they share the same representative vectors.

\paragraph{Micro experiments} are conducted to show how this approach works better than a random strategy.
We use 392 unlabeled samples from the same client on \texttt{MNLI}~\cite{williams2017broad}, and compare the filtered samples through our approach and a random one.
In Figure~\ref{fig:design-filter-diversity}, we visualize the distance between those samples by the cosine similarity on the Sentence-BERT output.
A key observation is that, our representativeness-aware filter can effectively select more diversified samples for pseudo labeling than a random filter.
We further compare the end-to-end performance in Figure~\ref{fig:design-filter-performance} with the same setting used in $\S$\ref{sec:eval}.
Our approach filters 95\% unlabeled samples and brings up to 7.2$\times$ convergence speedup on \texttt{MNLI}, with less than 1\% accuracy loss.
In comparison, randomly filtering data for pseudo labeling degrades the model accuracy significantly.

\subsection{Training Depth/Capacity Co-planning} \label{sec:design-optimization}

In order to enhance training efficiency in terms of aspects such as energy consumption per batch and memory usage, a prevailing method is to limit the \textit{depth} of the layers~\cite{guo2019spottune,lin-etal-2022-fednlp}. This technique, also referred to as \textit{freezing}, ensures that backward propagation is only performed across the top \texttt{k} layers nearest to the model output and encapsulate task-specific knowledge.
Another underutilized strategy pertains to controlling the \textit{capacity} of the layer, or determining how many weights to update within each layer. Recent research~\cite{zaken2021bitfit,logan2021cutting, cai2020tinytl} indicates that adjusting only the bias of a layer, which typically represents a meager 0.1\% of total weights, whilst maintaining the other weights as frozen, does not detrimentally affect model accuracy.
This is attributed to that bias values are responsible for the predicted class or hidden-state. Altering the bias allows for an efficient modification of the output vocabulary, thereby mitigating the risks of catastrophic forgetting or overfitting~\cite{french1999catastrophic, kemker2018measuring, ying2019overview, hawkins2004problem}. 

\begin{figure}[t]
	\centering
	\centering
    \begin{minipage}[b]{0.49\textwidth}
        \begin{minipage}[b]{0.49\textwidth}
            \includegraphics[width=1.01\textwidth]{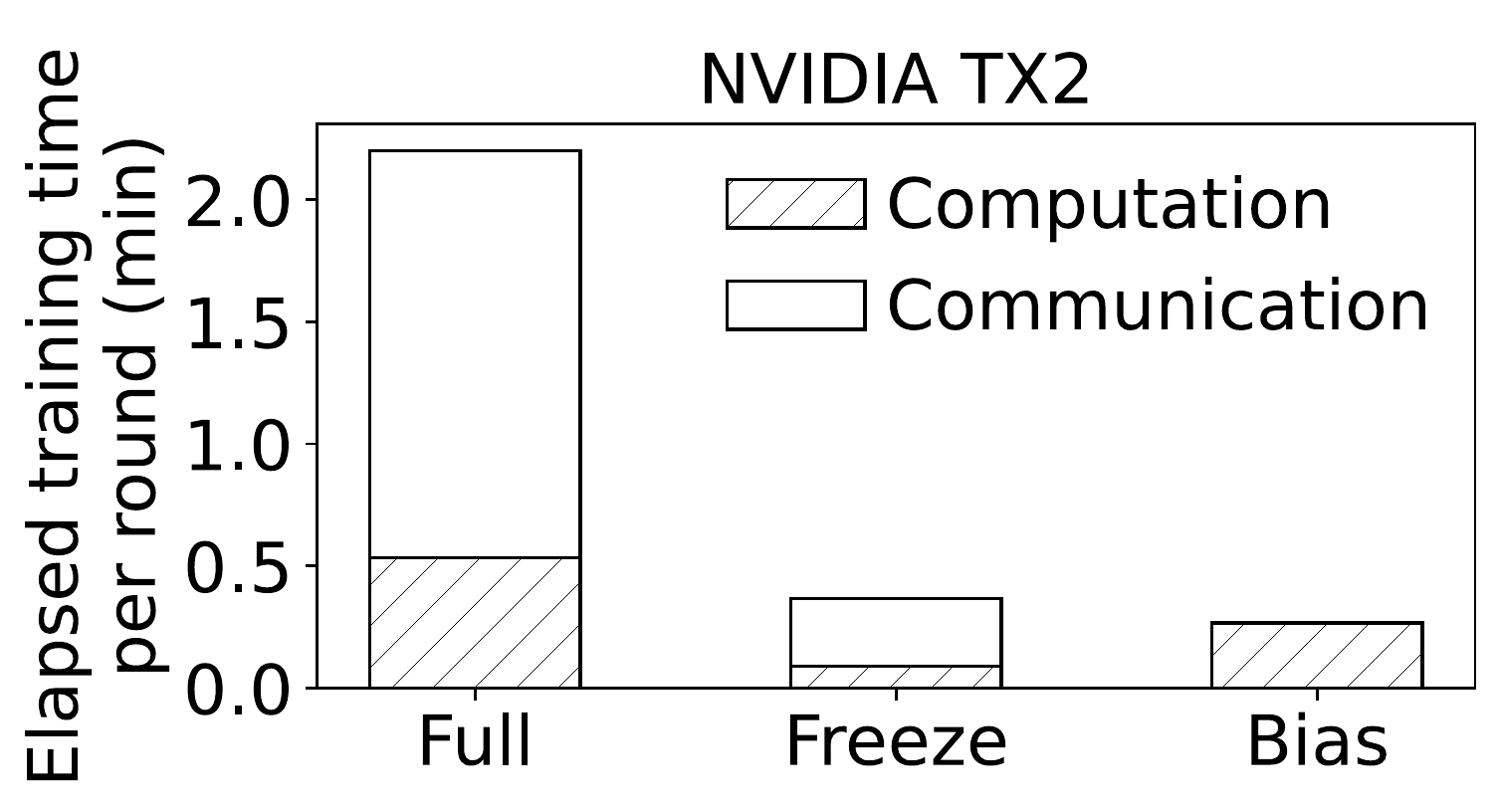}
        \end{minipage}
        ~
        \begin{minipage}[b]{0.49\textwidth}
            \includegraphics[width=0.9\textwidth]{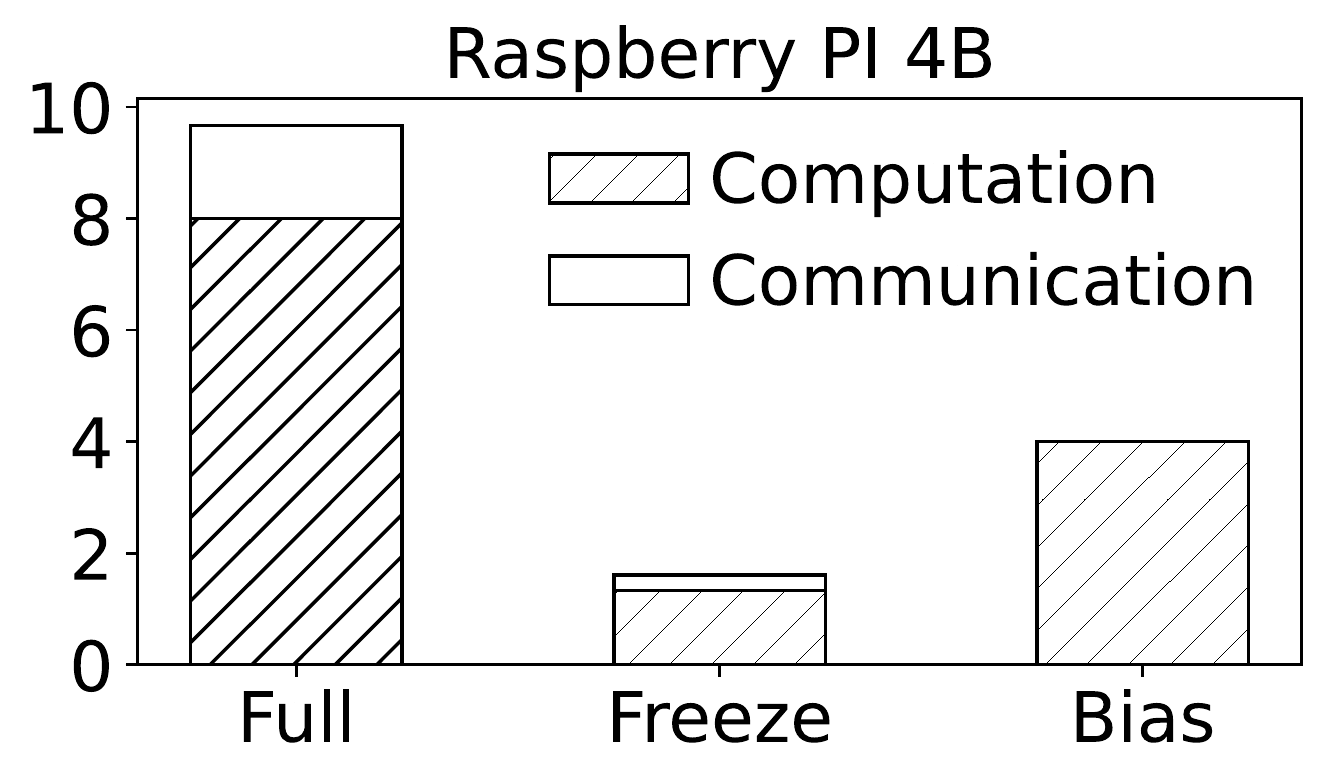}
        \end{minipage}
    \end{minipage}
	 \vspace{-25pt}
	\caption{The per-round compute and communication time with different on-device training optimizations. 
    Full: training everything; Freeze: freeze as many layers as possible with acceptable loss; Bias: training only bias for each layer.
        } 
	\vspace{-20pt}
	\label{fig:design-training}
\end{figure}
\textbf{Observation: compute-network tradeoff by controlling the layer depth and capacity.}
In the context of FedFSL, however, we observe that controlling either the layer depth or capacity alone is inadequate towards resource-efficient training on client devices.
Layer freezing can almost linearly scale down the resource cost with the freezed depth, e.g., up to 83\% in \texttt{AGNEWS} (20 out of 24 layers) without accuracy degradation.
However, the non-freezed layers still comprise of hundreds of MBs weights, in which case the network transmission bottlenecks the learning process.
Meanwhile, controlling the layer capacity by only updating model bias brings much more significant reduction on the communication;
yet the compute time bottlenecks as it still demands a complete forward-backward pass.

\begin{figure}[t]
	\centering
	 \includegraphics[width=0.48\textwidth]{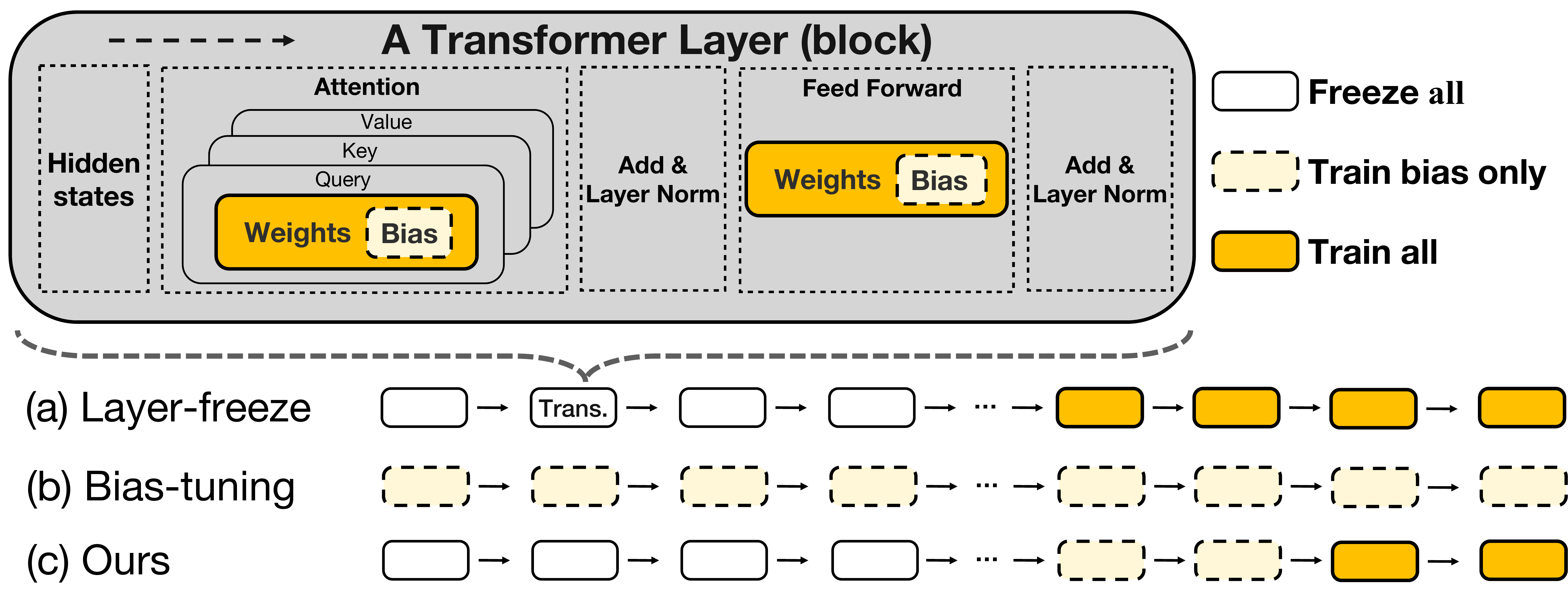}
	\vspace{-20pt}
	\caption{Comparing our approach of co-planning the tuned layer depth and capacity to traditional ones.}
	\vspace{-10pt}
	\label{fig:design-training-structure}
\end{figure}
Naive use of bias-tuning brings a significant runtime reduction on high-end mobile devices.
But on a much wimpy device, the speed-up drops. 
As shown in Figure~\ref*{fig:design-training}, on NVIDIA TX2, bias-tuning reduces the elapsed training time per FL round by 8.2$\times$.
Because TX2 is with strong GPU capacity, the compute cost is not heavy, leaving the network as the bottleneck, which is alleviated by bias-tuning.
While on Raspberry Pi 4B, bias-tuning's speed-up drops to 2.4$\times$, because RPI4B computes slow with weak CPU capacity, which is the bottleneck that vanilla bias-tuning could not handle.
In contrast, freezing the top layers can reduce the elapsed training time per round by 6.0$\times$ on RPI4B because it brings linear computation reduction.

\textbf{Co-planning the tuned layer depth and capacity}
To balance the compute and network, \sys carefully controls both the layer depth and capacity.
Simply applying both techniques, i.e., freezing some layers and updating the bias of the rest layers, does not fully exploit the potential of them.
Instead, we propose a mixed depth/capacity tuning paradigm shown in Figure~\ref{fig:design-training-structure}:
trains very few top layers with full capacity;
trains a few middle layers with reduced capacity (i.e., only bias updated);
freezes the other bottom layers.
The rationale of such a terraced design is that, the layer closer to the output encodes more downstream knowledge which can be extracted with limited data labels.
The concrete tuning decision is done offline on cloud.
Specifically, we use a FedFSL simulator that uses binary search to identify the optimal configuration.
The simulator takes three types of input:
pre-trained model;
datasets, which could be a public one with similar classification difficulty to the private dataset distributed across clients;
the estimated FedFSL runtime parameters including on-device training time and network bandwidth.

In comparison, a random scheme often results in low convergence accuracy or high resource costs due to either freezing too many or too few layers. For instance, when considering the \texttt{AGNEWS} dataset, compared to our cherry-picked plan (where we freeze 23 out of 24 layers, excluding only the bias of layers 16--23), freezing all 23 layers leads to a significant 4.1\% accuracy loss, while freezing just 16 layers incurs a slowdown of 13.8$\times$ to achieve only a marginal accuracy gain.

\paragraph{Integration with other parameter-efficient fine-tuning methods}
Enormous parameter-efficient fine-tuning methods are off-the-shelf to fine-tune large language models, such as adapter~\cite{pfeiffer2020adapterhub,cai2022autofednlp}, LoRA~\cite{he2021towards,hu2021lora}, etc.
Our system already includes one popular technique: bitfit~\cite{zaken2021bitfit}, that greatly saves tunable parameters while preserving the few-shot ability of large language models~\cite{logan2021cutting}. 
Other parameter-efficient fine-tuning methods could also interplay our co-planning schedule and reap benefits.
For instance, adapters could be utilized to fine-tune the middle layers with reduced capacity, while fine-tuning the very top layers with full capacity, which will not only lead to high parameter-efficiency but also high computation-efficiency.

    \section{Evaluation} 
\label{sec:eval}
We evaluate \sys to answer the following key questions: 
1) How much performance improvement (in terms of time-to-accuracy and relative model accuracy) does \sys achieve? 
2) How much performance improvement does \sys achieves across different number of gold labels?
3) How much performance improvement does each component of \sys contribute?
4) How much resource does \sys save?

\subsection{Implementation and Setup} \label{sec:eval-setup}



\begin{table}[]
    \resizebox{\columnwidth}{!}{%
    \begin{tabular}{c|cccc}
    \hline
    Dataset    & \multicolumn{1}{c|}{\texttt{AGNEWS}~\cite{zhang2015character}} & \multicolumn{1}{c|}{\texttt{MNLI}~\cite{williams2017broad}} & \multicolumn{1}{c|}{\texttt{YAHOO}~\cite{zhang2015character}} & \texttt{YELP-F}~\cite{zhang2015character} \\ \hline
    \# Training  & \multicolumn{1}{c|}{120k}   & \multicolumn{1}{c|}{392.7k}  & \multicolumn{1}{c|}{1.4M}   & 650k   \\
    \# Test    & \multicolumn{1}{c|}{7.6k}   & \multicolumn{1}{c|}{9.8k} & \multicolumn{1}{c|}{60k}   & 50k  \\
    \# Clients & \multicolumn{1}{c|}{100}    & \multicolumn{1}{c|}{1000} & \multicolumn{1}{c|}{1000}  & 1000 \\ \hline
    \# Labels  & \multicolumn{1}{c|}{64}     & \multicolumn{1}{c|}{64}   & \multicolumn{1}{c|}{64}    & 64   \\
    Distribution & \multicolumn{1}{c|}{Skewed} & \multicolumn{1}{c|}{Uniform} & \multicolumn{1}{c|}{Skewed} & Skewed \\ \hline
    Prompt     &  \multicolumn{1}{c|}{a  \_\_\_\_  b}       &  \multicolumn{1}{c|}{a ?\_\_\_\_, b}    &  \multicolumn{1}{c|}{Category: a \_\_\_\_ b}      & It was \_\_\_\_. a     \\ \hline
    \end{tabular}%
    }
    \caption{Evaluation datasets.
    Label quantity of each class follows prior work~\cite{lin-etal-2022-fednlp} where $\alpha=1$.
    Please note that 64 is the total number of labels across clients, not per client.}
    \label{tab:eval-datasets}
    \vspace{-25pt}
    \end{table}

\textbf{\sys prototype}
We have fully implemented the \sys prototype atop \texttt{PET} \cite{schick2020exploiting} and FedNLP~\cite{lin-etal-2022-fednlp}.
PET is a popular prompt learning framework for NLP tasks.
FedNLP is the state-of-the-art framework for evaluating NLP tasks under federated setting.
As prior work~\cite{bonawitz2019towards1}, we adopt the parameter server (PS) architecture among the clients and central server.
The on-device training and inference performance is tested with PyTorch 1.10, and then plugged into FedNLP framework.
The models trained through prompt learning will be collected in the central server and aggregated through \textit{FedAvg}~\cite{mcmahan2017communication} algorithm, which is also the default setting in prior FedNLP literature~\cite{lin-etal-2022-fednlp}.
Both pseudo labeling and prompt learning randomly select clients for labeling and training per round.

\begin{table}[]
    \resizebox{\columnwidth}{!}{%
    \begin{tabular}{c|cc|cc}
    \hline
    \multirow{2}{*}{\textbf{Setup}} & \multicolumn{2}{c|}{\textbf{Labeling}}  & \multicolumn{2}{c}{\textbf{Training}}                                                   \\ \cline{2-5} 
                                    & \textbf{Pacing} & \textbf{Optimization} & \textbf{Method} & \textbf{Optimization}                                                 \\ \hline
    \texttt{FedCLS}      & /      & / & Head-based   & /                \\ \hline
    \texttt{FedFSL}      & Static & / & Prompt-based & /                \\ \hline
    \texttt{FedFSL-BIAS} & Static & / & Prompt-based & Bias-only tuning \\ \hline
    \texttt{\sys (Ours)}                            & \begin{tabular}[c]{@{}c@{}}Curriculum \\ ($\S$\ref{sec:design-pacing})\end{tabular}      & \begin{tabular}[c]{@{}c@{}}Filtering \\ ($\S$\ref{sec:design-filter})\end{tabular}              & \begin{tabular}[c]{@{}c@{}}Prompt-based \\ ($\S$\ref{sec:algo-blocks})\end{tabular}    & \begin{tabular}[c]{@{}c@{}}Depth/Capacity \\ Co-planning ($\S$\ref{sec:design-optimization})\end{tabular} \\ \hline
    \end{tabular}%
    }
    \caption{Summary of baselines used in experiments.}
    \label{tab:eval-baseline}
    \vspace{-10pt}
    \end{table}

\noindent \textbf{Baselines} 
We compare \sys to the following alternatives and the key differences are summarized in Table~\ref{tab:eval-baseline}.
(1) \texttt{FedCLS} is the vanilla federated fine-tuning method without optimizations~\cite{devlin2018bert,sanh2019distilbert}.
It trains only with the limited gold labels.
(2) \texttt{FedFSL} implements pseudo labeling and prompt learning but without our system optimizations.
(3) \texttt{FedFSL-BIAS}
runs the \texttt{FedFSL} pipeline; it however tunes only the layer bias while freezing the other weights.
Both \texttt{FedFSL} and \texttt{FedFSL-BIAS} use static pacing, i.e., adding 100 pseudo labels per selected client per round.
\sys and all baselines use the same set of hyper-parameters as prior literature~\cite{lin-etal-2022-fednlp, schick2020exploiting,cai2022autofednlp}: mini-batch size as 4; local training epoch as 1; max sequence length as 256, per-round participant client number as 5.


\noindent \textbf{Models}
We test two foundation NLP models: RoBERTa-large~\cite{liu2019roberta} (default) and its light version RoBERTa-base, composed of 24 and 12 transformer layers, respectively.
They are downloaded directly from Huggingface~\cite{wolf2019huggingface}. 
We use RoBERTa-large for most of our experiments for its superior accuracy performance, as we discussed in $\S$\ref{sec:bkgnd-fednlp}.
Generative tasks requiring mega large language models like GPT3~\cite{brown2020language} are out of this work's scope. 
For most classification or seq2seq tasks, BERT-like models are adequate to achieve usable accuracy. 

\noindent \textbf{Dataset and few-shot setting} 
We experiment with four popular NLP datasets and prompts\footnote{We attempt 6, 2, 6, 4 different prompts that are widely used in prior work for each dataset, and use the one with highest accuracy.
The verbalizers are the same as the previous literature~\cite{schick2020exploiting}.},
as shown in Table~\ref{tab:eval-datasets}.
(1) \texttt{AGNEWS}~\cite{zhang2015character} is a news classification dataset. 
(2) \texttt{MNLI}~\cite{williams2017broad} is a sentence understanding dataset. 
(3) \texttt{YELP Review Full (YELP-F)}~\cite{zhang2015character} is a restaurant rating dataset. 
(4) \texttt{YAHOO}~\cite{zhang2015character} is a question-answer pairing dataset.
For each dataset, we follow prior work~\cite{cai2022autofednlp,fan2021federated} to randomly select gold labels.
By default, the labels form a skewed distribution across clients to be more realistic to real-world situation as discussed in $\S$\ref{sec:bkgnd-fednlp}.
For each dataset, we generate 3 different few-shot settings, on which we repeat the experiments and report the mean results.

\noindent \textbf{Hardware}
As prior FL literature~\cite{lin-etal-2022-fednlp,li2021hermes,lipyramidfl,lai2020oort,shin2022fedbalancer}, our experiments are carried out in an emulation manner on a GPU server with 8x NVIDIA A40.
The on-device training time is obtained on 2 development boards with similar hardware capacity to mainstream mobile devices, i.e., NVIDIA TX2~\cite{tx2} and Raspberry Pi 4B~\cite{rpi4b}
The default testbed device is NVIDIA TX2 without special statement. 
The numbers are then plugged into the emulation framework to calculate the elapsed time.
We try various network bandwidths between clients and server, with default number as 1MB/s, a typical setting for mobile and IoT devices~\cite{wifi-state,han2016mp}.

\subsection{End-to-end Performance} \label{sec:eval-overall}
\begin{figure*}[t]
    \centering
    \begin{minipage}[b]{1\textwidth}
        \begin{minipage}[b]{0.24\textwidth}
            \includegraphics[width=1\textwidth]{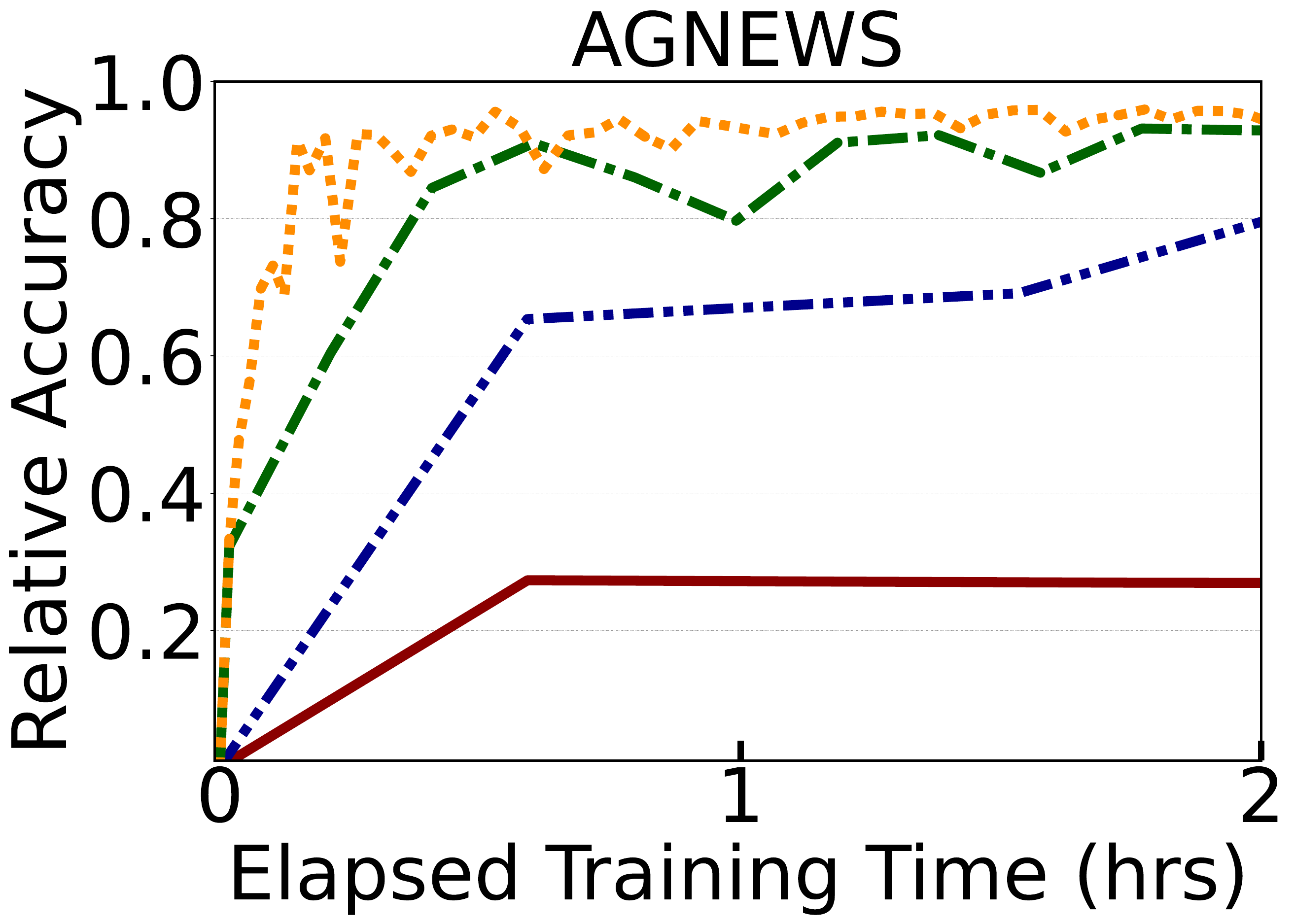}
        \end{minipage}
        ~
        \begin{minipage}[b]{0.24\textwidth}
            \includegraphics[width=1.875\textwidth]{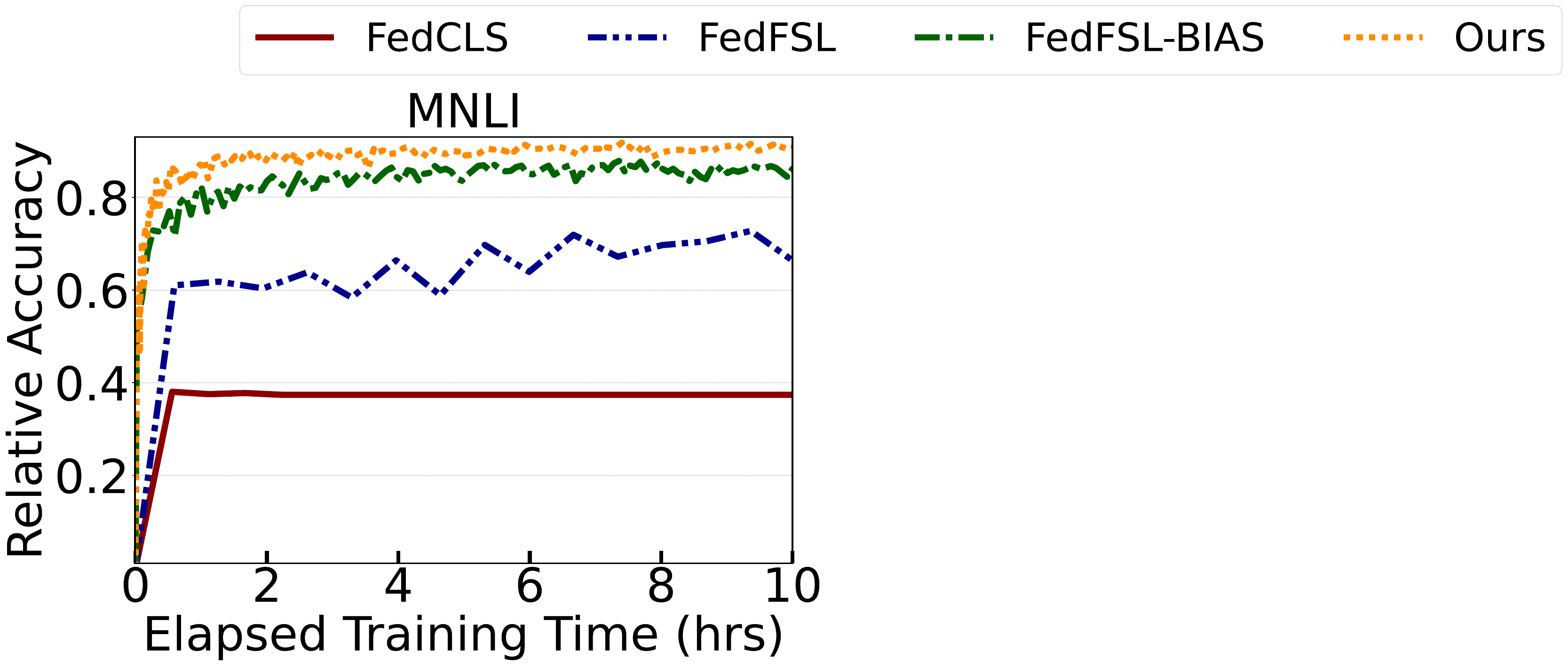}
        \end{minipage}
        ~
        \begin{minipage}[b]{0.24\textwidth}
            \includegraphics[width=1\textwidth]{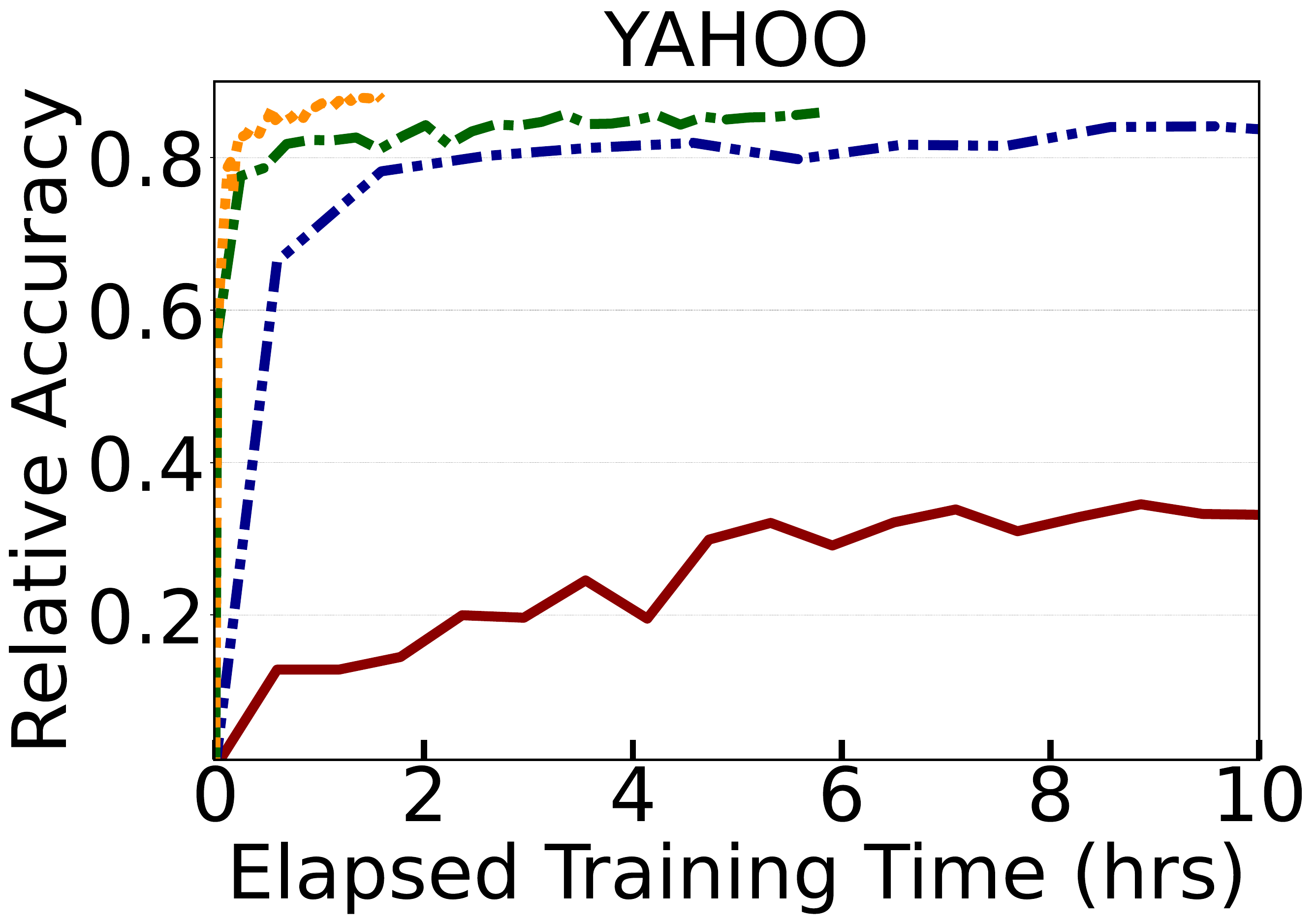}
        \end{minipage}
        ~
        \begin{minipage}[b]{0.238\textwidth}
            \includegraphics[width=1\textwidth]{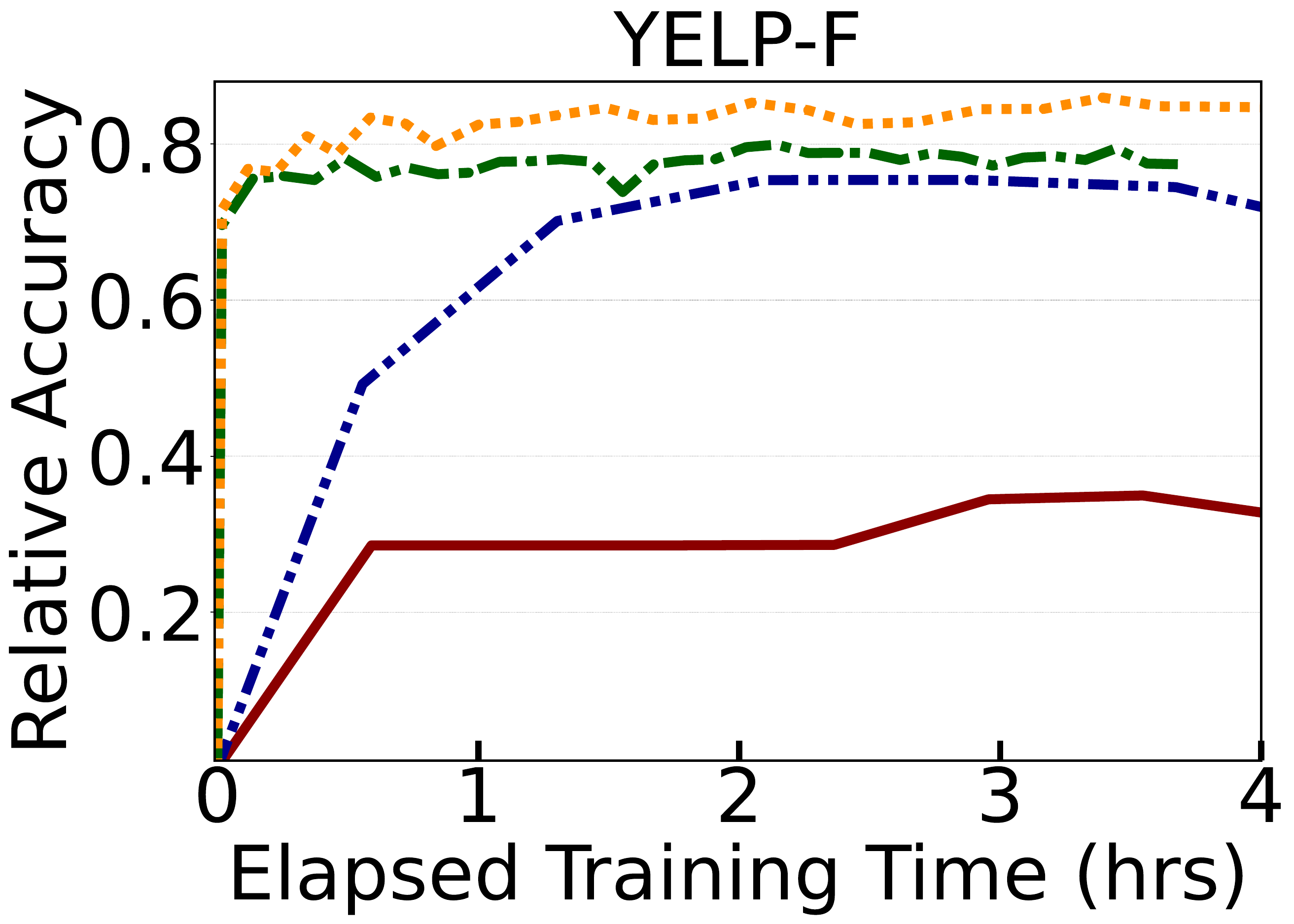}
        \end{minipage}
        \vspace{-5pt}
        \subcaption{RoBERTa-large~\cite{liu2019roberta}}
        \label{fig:eval-overall-roberta-large}
    \end{minipage}


    \begin{minipage}[b]{1\textwidth}
        \begin{minipage}[b]{0.24\textwidth}
            \includegraphics[width=1\textwidth]{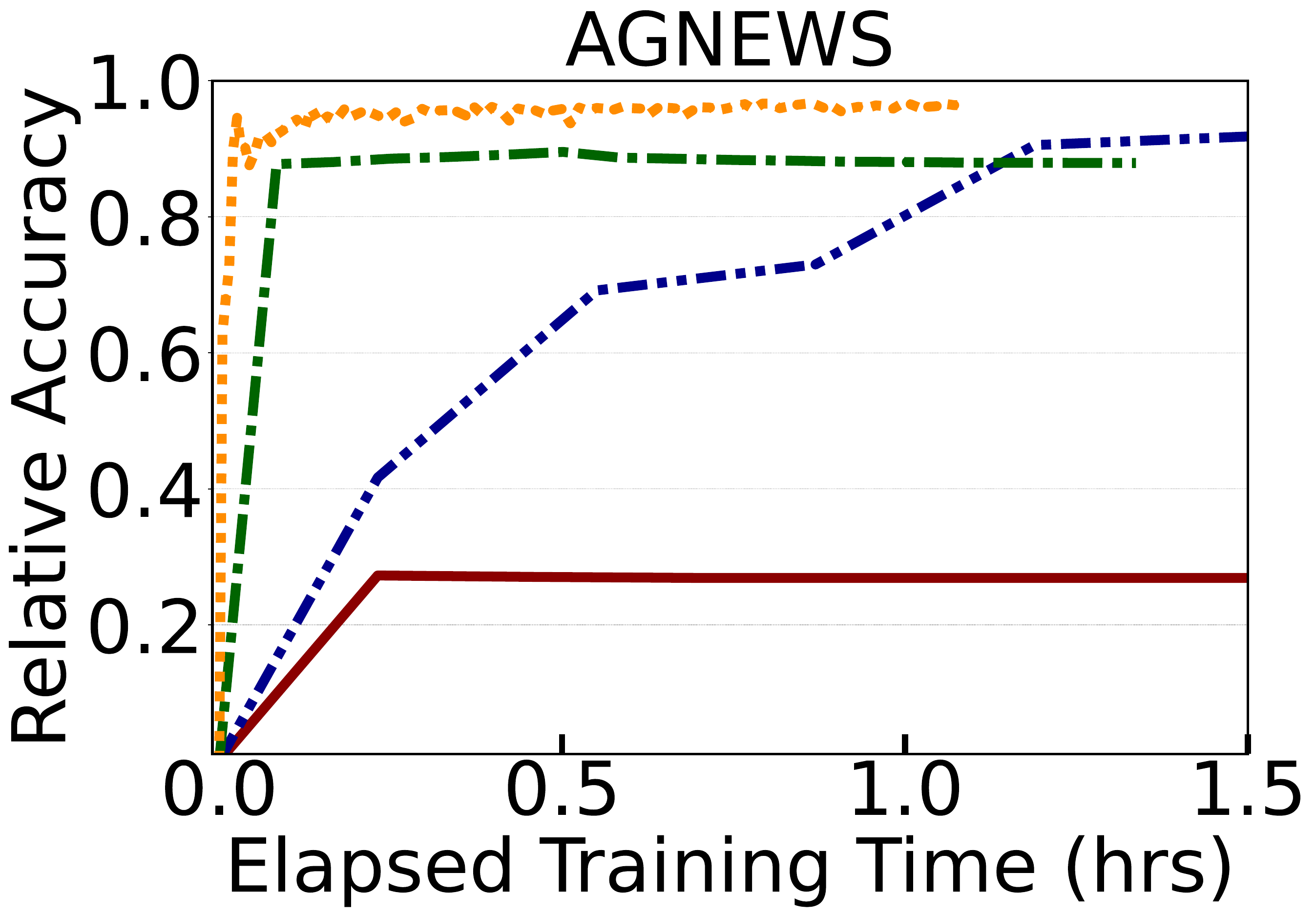}
        \end{minipage}
        ~
        \begin{minipage}[b]{0.24\textwidth}
            \includegraphics[width=1\textwidth]{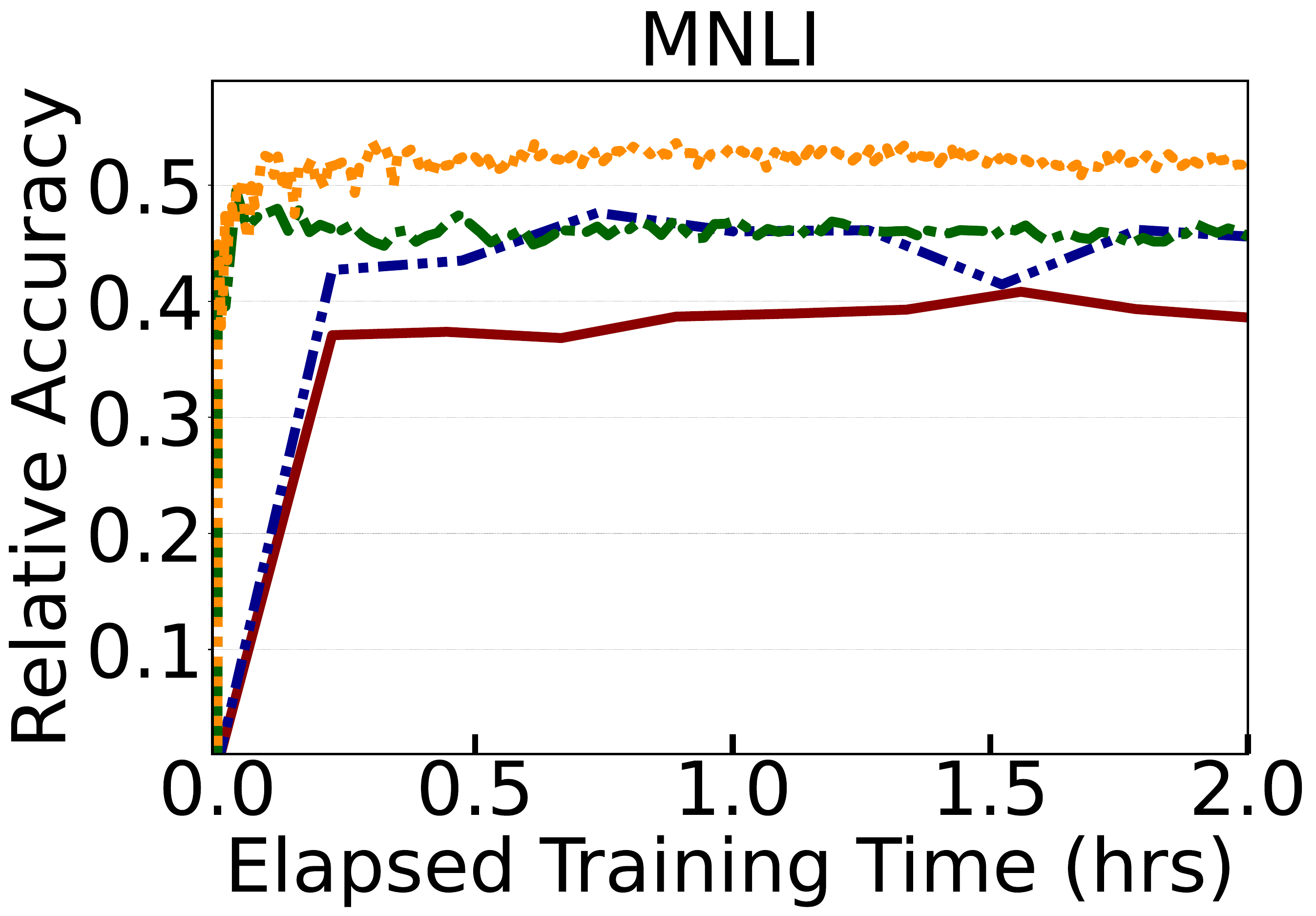}
        \end{minipage}
        ~
        \begin{minipage}[b]{0.235\textwidth}
            \includegraphics[width=1\textwidth]{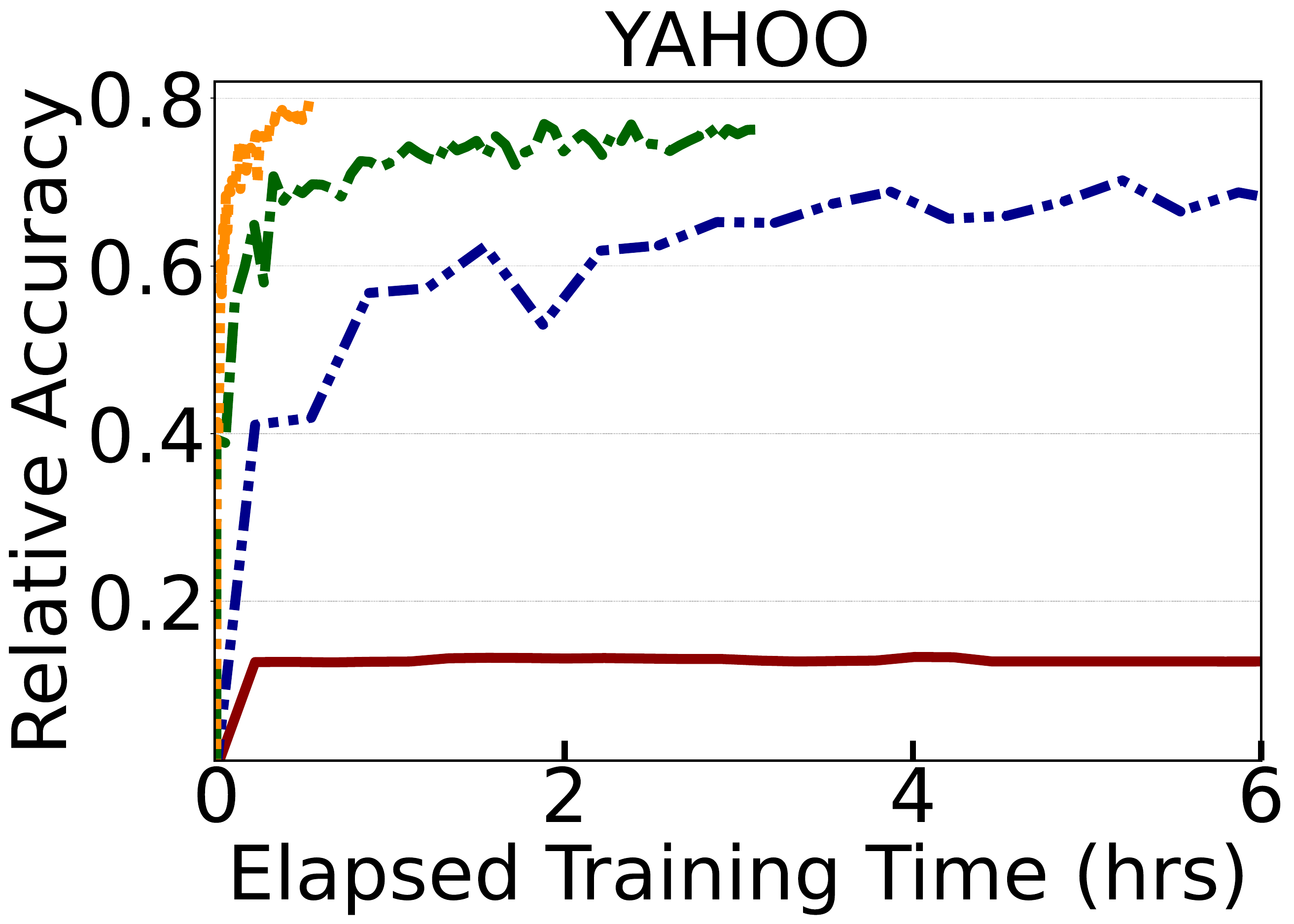}
        \end{minipage}
        ~
        \begin{minipage}[b]{0.24\textwidth}
            \includegraphics[width=1\textwidth]{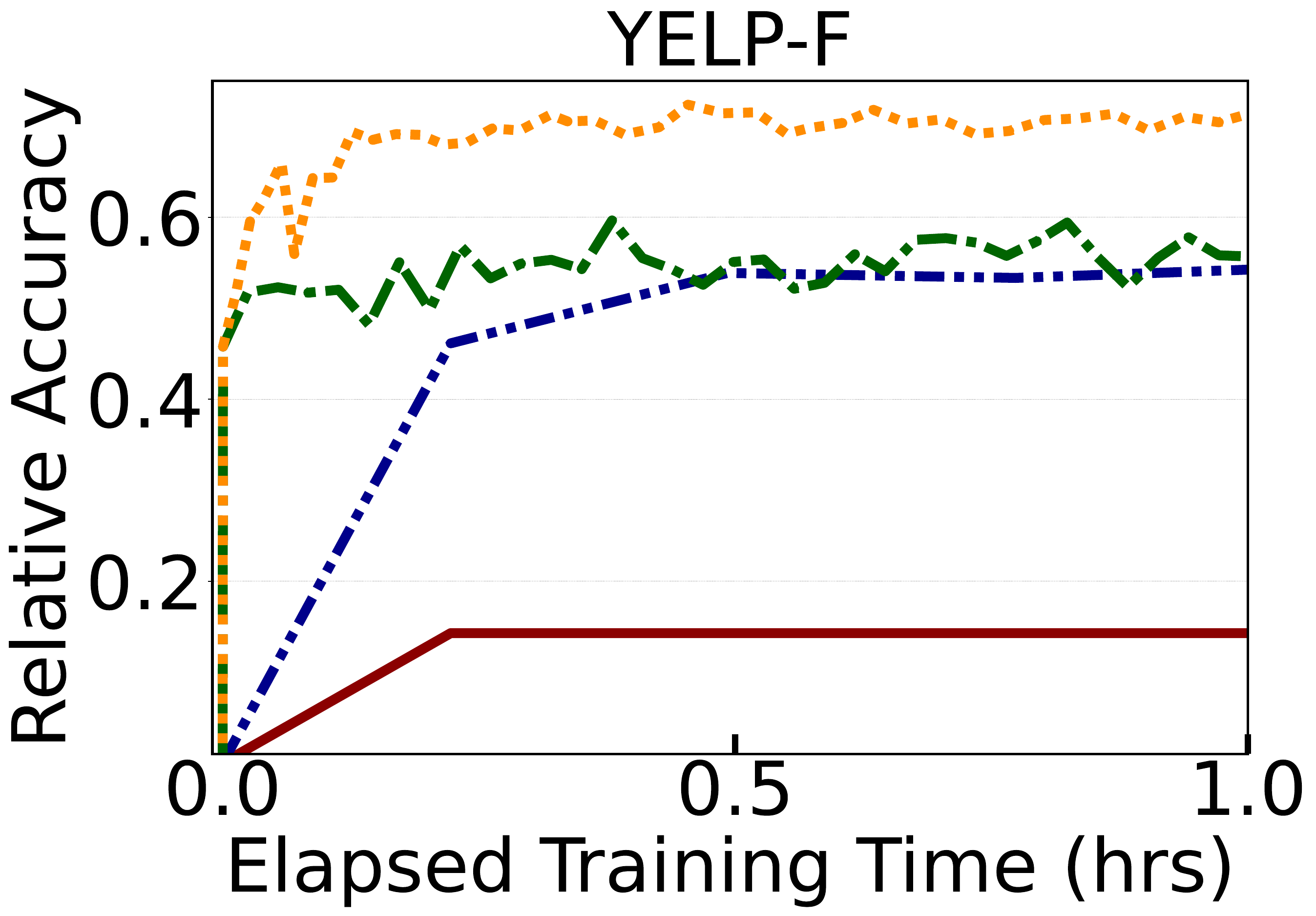}
        \end{minipage}
        \vspace{-5pt}
        \subcaption{RoBERTa-base~\cite{liu2019roberta}} \label{fig:eval-overall-roberta-base}
        \vspace{-10pt}
    \end{minipage}
    \vspace{-15pt}
    \caption{Overall Performance of \sys and baselines.
    Device: Jetson TX2.
    }
    
    \vspace{-5pt}
\end{figure*}

\begin{table*}[t]
\resizebox{2.1\columnwidth}{!}{%
\begin{tabular}{c|ccccc|ccccc|ccccc|ccccc}
\hline
\textbf{Dataset} &
  \multicolumn{5}{c|}{\textbf{AGNEWS}} &
  \multicolumn{5}{c|}{\textbf{MNLI}} &
  \multicolumn{5}{c|}{\textbf{YAHOO}} &
  \multicolumn{5}{c}{\textbf{YELP-F}} \\ \hline
 &
  \multicolumn{1}{c|}{} &
  \multicolumn{4}{c|}{Time-to-acc (hr)} &
  \multicolumn{1}{c|}{} &
  \multicolumn{4}{c|}{Time-to-acc (hr)} &
  \multicolumn{1}{c|}{} &
  \multicolumn{4}{c|}{Time-to-acc (hr)} &
  \multicolumn{1}{c|}{} &
  \multicolumn{4}{c}{Time-to-acc (hr)} \\ \cline{3-6} \cline{8-11} \cline{13-16} \cline{18-21} 
 &
  \multicolumn{1}{c|}{} &
  \multicolumn{2}{c|}{TX2} &
  \multicolumn{2}{c|}{RPI} &
  \multicolumn{1}{c|}{} &
  \multicolumn{2}{c|}{TX2} &
  \multicolumn{2}{c|}{RPI} &
  \multicolumn{1}{c|}{} &
  \multicolumn{2}{c|}{TX2} &
  \multicolumn{2}{c|}{RPI} &
  \multicolumn{1}{c|}{} &
  \multicolumn{2}{c|}{TX2} &
  \multicolumn{2}{c}{RPI} \\ \cline{3-6} \cline{8-11} \cline{13-16} \cline{18-21} 
\multirow{-3}{*}{\textbf{Perf.}} &
  \multicolumn{1}{c|}{\multirow{-3}{*}{\begin{tabular}[c]{@{}c@{}}Conv.\\ Acc.\end{tabular}}} &
  \multicolumn{1}{c|}{acc1} &
  \multicolumn{1}{c|}{acc2} &
  \multicolumn{1}{c|}{acc1} &
  acc2 &
  \multicolumn{1}{c|}{\multirow{-3}{*}{\begin{tabular}[c]{@{}c@{}}Conv.\\ Acc.\end{tabular}}} &
  \multicolumn{1}{c|}{acc1} &
  \multicolumn{1}{c|}{acc2} &
  \multicolumn{1}{c|}{acc1} &
  acc2 &
  \multicolumn{1}{c|}{\multirow{-3}{*}{\begin{tabular}[c]{@{}c@{}}Conv.\\ Acc.\end{tabular}}} &
  \multicolumn{1}{c|}{acc1} &
  \multicolumn{1}{c|}{acc2} &
  \multicolumn{1}{c|}{acc1} &
  acc2 &
  \multicolumn{1}{c|}{\multirow{-3}{*}{\begin{tabular}[c]{@{}c@{}}Conv.\\ Acc.\end{tabular}}} &
  \multicolumn{1}{c|}{acc1} &
  \multicolumn{1}{c|}{acc2} &
  \multicolumn{1}{c|}{acc1} &
  acc2 \\ \hline
\textbf{FedCSL} &
  \multicolumn{1}{c|}{27.9\%} &
  X &
  X &
  X &
  X &
  \multicolumn{1}{c|}{37.3\%} &
  X &
  X &
  X &
  X &
  \multicolumn{1}{c|}{34.6\%} &
  X &
  X &
  X &
  X &
  \multicolumn{1}{c|}{35.7\%} &
  X &
  X &
  X &
  X \\ \hline
\textbf{FedFSL} &
  \multicolumn{1}{c|}{92.5\%} &
  3.3 &
  3.3 &
  50.0 &
  50.0 &
  \multicolumn{1}{c|}{74.1\%} &
  9.2 &
  X &
  137.5 &
  X &
  \multicolumn{1}{c|}{84.3\%} &
  8.3 &
  X &
  125.0 &
  X &
  \multicolumn{1}{c|}{75.3\%} &
  2.1 &
  X &
  31.3 &
  X \\ \hline
\textbf{FedFSL-BIAS} &
  \multicolumn{1}{c|}{92.5\%} &
  1.7 &
  1.7 &
  25.0 &
  25.0 &
  \multicolumn{1}{c|}{88.1\%} &
  0.5 &
  11.7 &
  7.5 &
  175.0 &
  \multicolumn{1}{c|}{85.9\%} &
  3.3 &
  5.3 &
  50.0 &
  80.0 &
  \multicolumn{1}{c|}{79.4\%} &
  0.2 &
  2.1 &
  2.5 &
  10.4 \\ \hline
\textbf{Ours} &
  \multicolumn{1}{c|}{{\color[HTML]{FD6864} \textbf{95.9\%}}} &
  {\color[HTML]{FD6864} \textbf{0.4}} &
  {\color[HTML]{FD6864} \textbf{0.4}} &
  {\color[HTML]{FD6864} \textbf{5.5}} &
  {\color[HTML]{FD6864} \textbf{5.5}} &
  \multicolumn{1}{c|}{{\color[HTML]{FD6864} \textbf{92.2\%}}} &
  {\color[HTML]{FD6864} \textbf{0.2}} &
  {\color[HTML]{FD6864} \textbf{0.8}} &
  {\color[HTML]{FD6864} \textbf{2.5}} &
  {\color[HTML]{FD6864} \textbf{12.5}} &
  \multicolumn{1}{c|}{{\color[HTML]{FD6864} \textbf{88.5\%}}} &
  {\color[HTML]{FD6864} \textbf{0.3}} &
  {\color[HTML]{FD6864} \textbf{0.7}} &
  {\color[HTML]{FD6864} \textbf{5.0}} &
  {\color[HTML]{FD6864} \textbf{10.0}} &
  \multicolumn{1}{c|}{{\color[HTML]{FD6864} \textbf{86.8\%}}} &
  {\color[HTML]{FD6864} \textbf{0.1}} &
  {\color[HTML]{FD6864} \textbf{0.5}} &
  {\color[HTML]{FD6864} \textbf{1.3}} &
  {\color[HTML]{FD6864} \textbf{7.5}} \\ \hline
\end{tabular}%
}
\caption{The final convergence accuracy (``Conv. Acc.'') and the elapsed training time (``Time-to-acc'') to reach different relative accuracy.
  NLP model: RoBERT-large.
  ``acc1''/``acc2'' are the final convergence accuracy of \texttt{FedFSL}/\texttt{FedFSL-BIAS}, respectively.
  ``X'' means the accuracy cannot be achieved.}
  \label{tab:eval-overall}
  \vspace{-20pt}
\end{table*}

\textbf{\sys significantly speeds up model convergence at high accuracy.}
As demonstrates in Figure~\ref{fig:eval-overall-roberta-large} and Table~\ref{tab:eval-overall}, \sys achieves 86.8\%-95.9\% relative accuracy to a full-label fine-tuning on four datasets, using only 64 data labels.
In comparison,  \texttt{FedCLS} only converges at 27.9\%--37.3\% relative accuracy. 
The improved performance is attributed to the pseudo labeling and prompt learning techniques adopted by \sys.
Accordingly, the other two baselines also achieve a competitive convergence accuracy.
However, \sys is much faster to converge:
\sys takes only 0.1--0.4 hours to reach the convergence accuracy of \texttt{FedFSL} on Jetson TX2, which is 8.2$\times$--46.0$\times$ faster than \texttt{FedFSL}.
Even when compared to the stronger baseline, \texttt{FedFSL-BIAS}, \sys still converges 4.3$\times$--14.6$\times$ faster.
On a wimpier device RPI 4B, \sys converges a few times slower due to the lengthened local training time.
Nevertheless, it consistently outperforms \texttt{FedFSL} by 28.3$\times$ and \texttt{FedFSL-BIAS} by 10.8$\times$ on average.


We also observe \sys achieves 3.4\%--18.1\% higher relative accuracy than \texttt{FedFSL} and \texttt{FedFSL-BIAS}, which are built atop the same algorithmic foundation.
There are two potential reasons.
First, by tuning fewer weights (bias), the training algorithm can better extract consolidated knowledge through fewer data labels.
Second, \sys employs curriculum pacing that effectively orchestrates pseudo labeling with federated learning,
whereas the two \texttt{FedFSL} baselines use a static pacing strategy that could lead to insufficient or excessive pseudo labels.

We then extend our experiments to RoBERTa-base and present the results in Figure \ref{fig:eval-overall-roberta-base}.
\sys achieves 8.0\%--66.9\% higher relative accuracy than \texttt{FedCLS}.
Compared to the more competitive baselines, i.e. \texttt{FedFSL}/\texttt{FedFSL-BIAS}, \sys maintains high convergence accuracy and is 24.1$\times$/4.6$\times$ faster on average, respectively.
In most cases, except for \texttt{AGNEWS}, which is a relatively easy task, using small models with weaker knowledge results in lower accuracy, as discussed in $\S$\ref{sec:bkgnd}.
For instance, RoBERTa-base sees a decrease of 38.5\%, 8.2\% and 13.9\% in convergence accuracy on \texttt{MNLI}, \texttt{YAHOO} and \texttt{YELP-F}, respectively, when compared to RoBERTa-large.
Fortunately, \sys significantly reduces the large model fine-tuning cost while maintaining its high performance.

\begin{figure}[t]
	\centering
	\vspace*{-4pt}\hspace*{3pt}\begin{minipage}[b]{0.24\textwidth}
        \hspace*{-6pt}\includegraphics[width=1.7\textwidth]{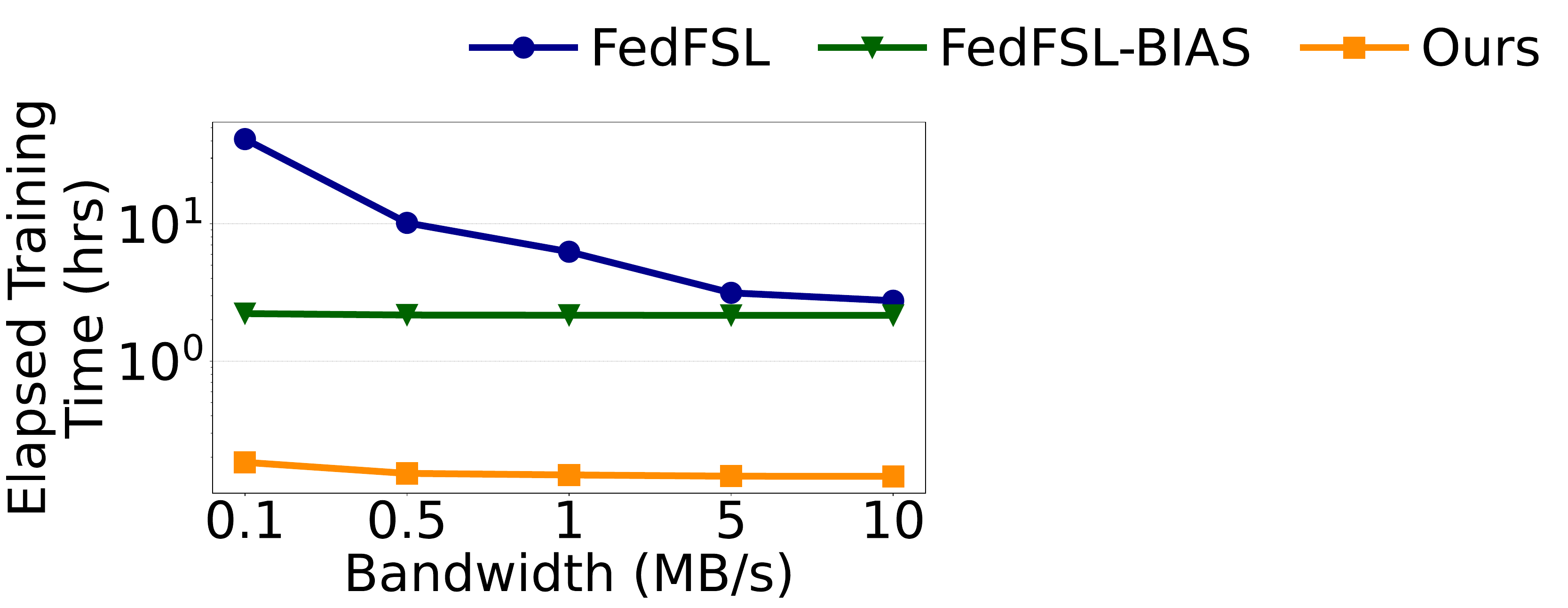}\vspace*{-4pt}
        \subcaption{\texttt{AGNEWS}}
    \end{minipage}
	~
    \hspace*{-14pt}\begin{minipage}[b]{0.24\textwidth}
        \hspace*{10pt}\includegraphics[width=0.88\textwidth]{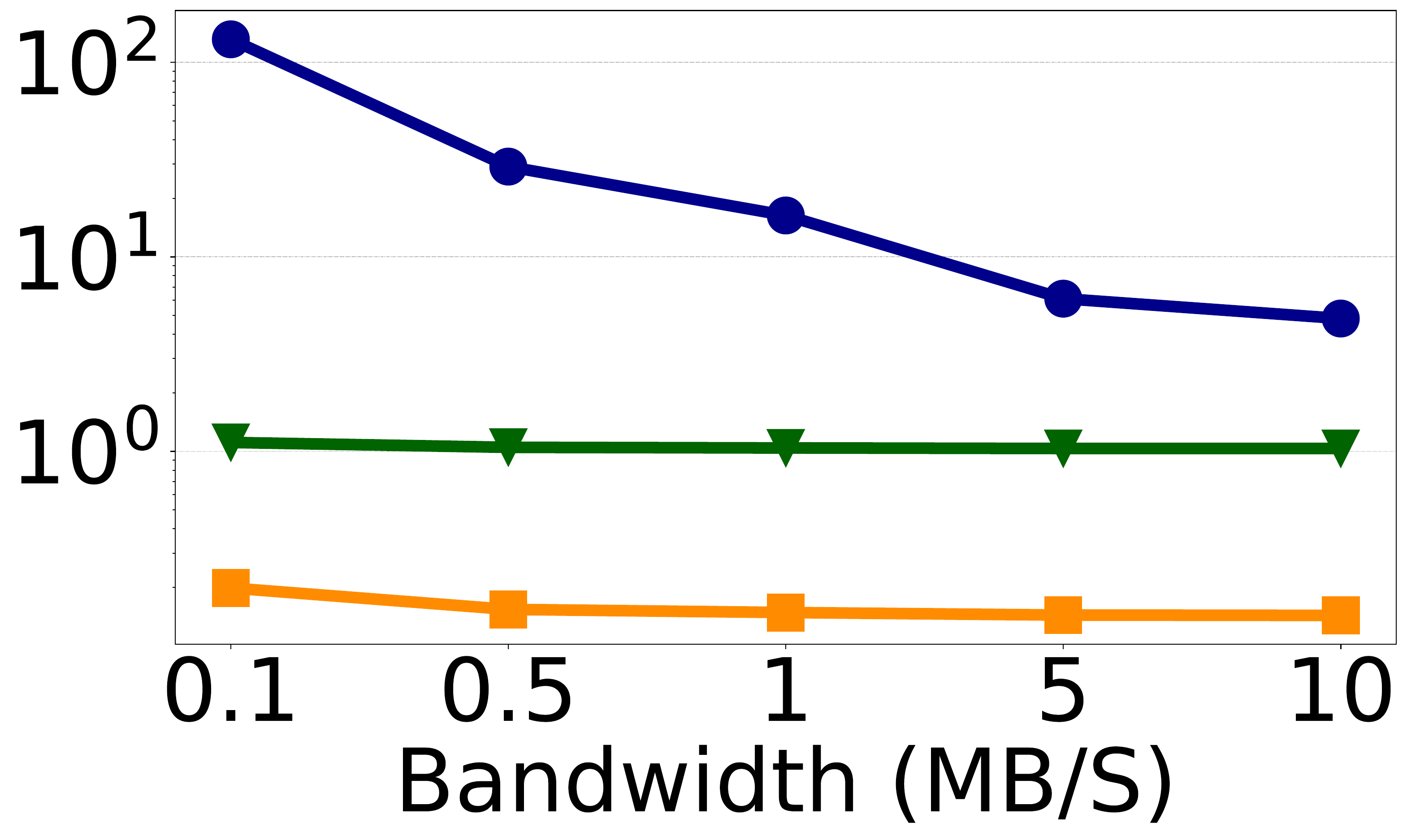}\vspace*{-2.5pt}
        \subcaption{\texttt{MNLI}}
    \end{minipage}
\vspace{-10pt}
	\caption{
	\sys{} outperforms baselines under all network bandwidths to reach the convergence accuracy of \texttt{FedFSL}.
	}
	\label{fig:eval-overall-bandwidth}
\end{figure}
\begin{figure*}[t]
    \centering
    \begin{minipage}[b]{0.49\textwidth}
        \begin{minipage}[b]{0.49\textwidth}
            \includegraphics[width=0.9\textwidth]{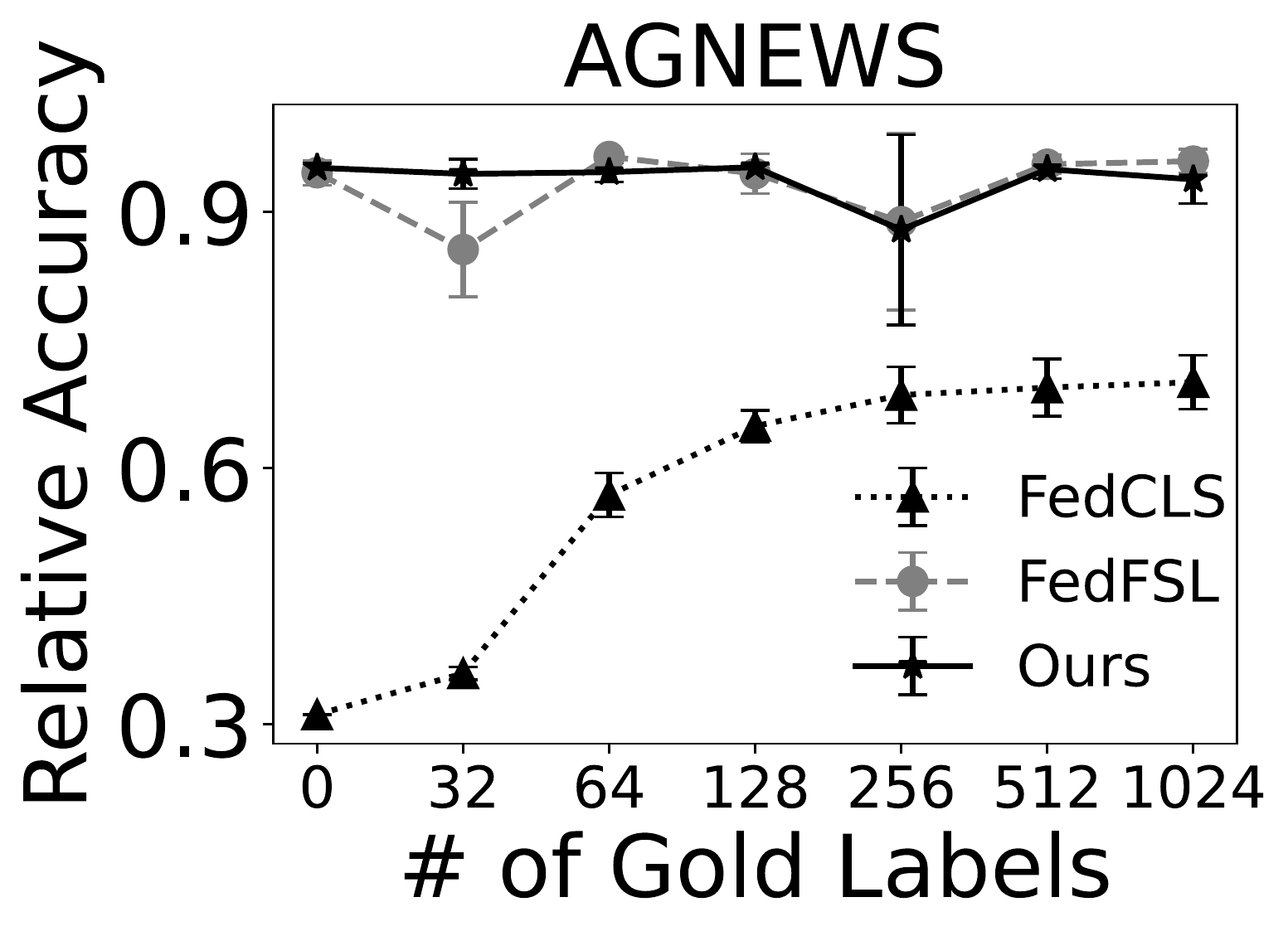}
        \end{minipage}
        ~
        \begin{minipage}[b]{0.49\textwidth}
            \includegraphics[width=0.9\textwidth]{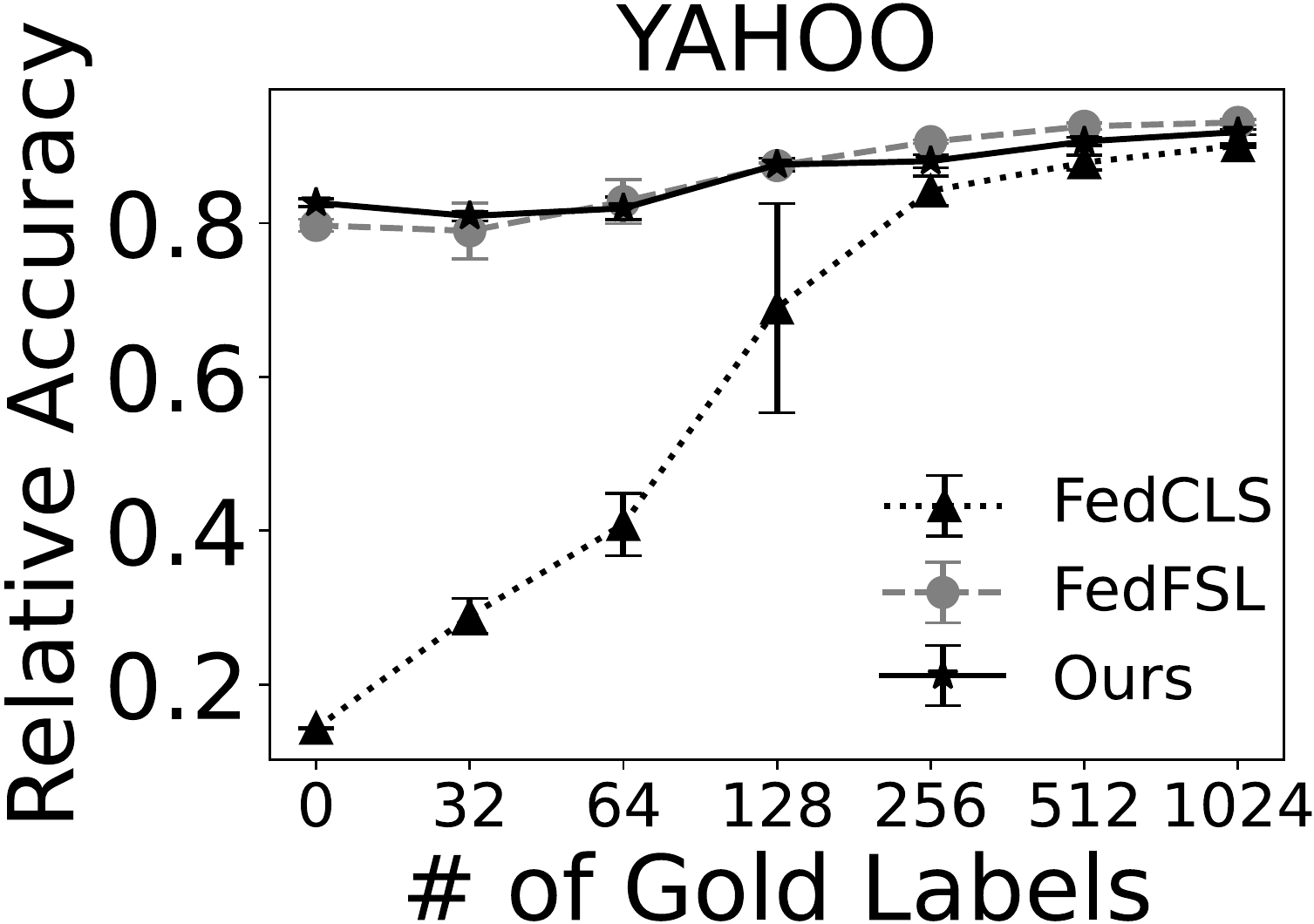}
        \end{minipage}
        \vspace{-5pt}
        \subcaption{Convergence accuracy}\label{fig:eval-overall-datapoints-acc}
    \end{minipage}
    \begin{minipage}[b]{0.49\textwidth}
        \begin{minipage}[b]{0.49\textwidth}
            \includegraphics[width=1.\textwidth]{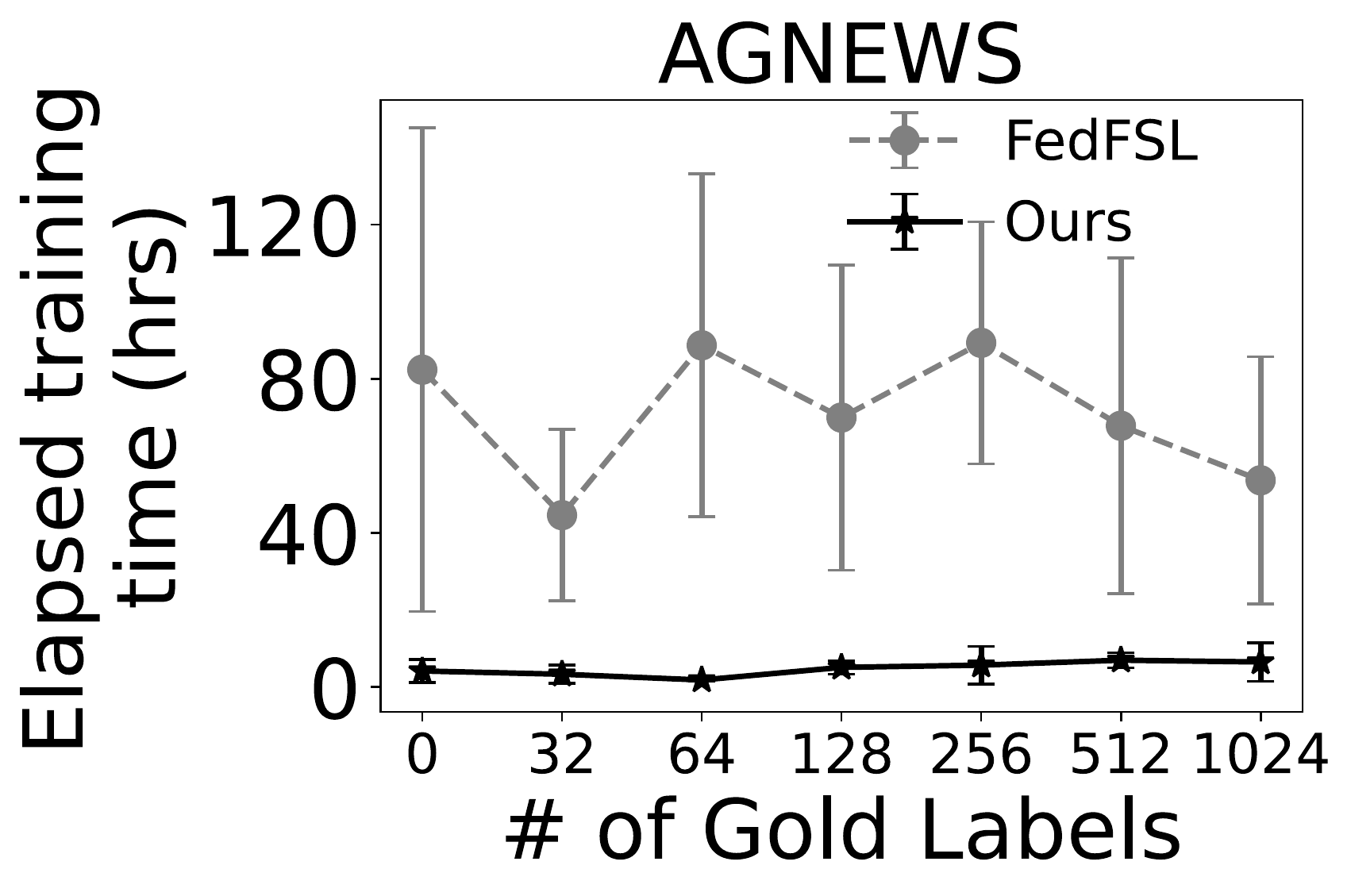}\vspace{-1.1pt}
        \end{minipage}
        ~
        \begin{minipage}[b]{0.49\textwidth}
            \includegraphics[width=1\textwidth]{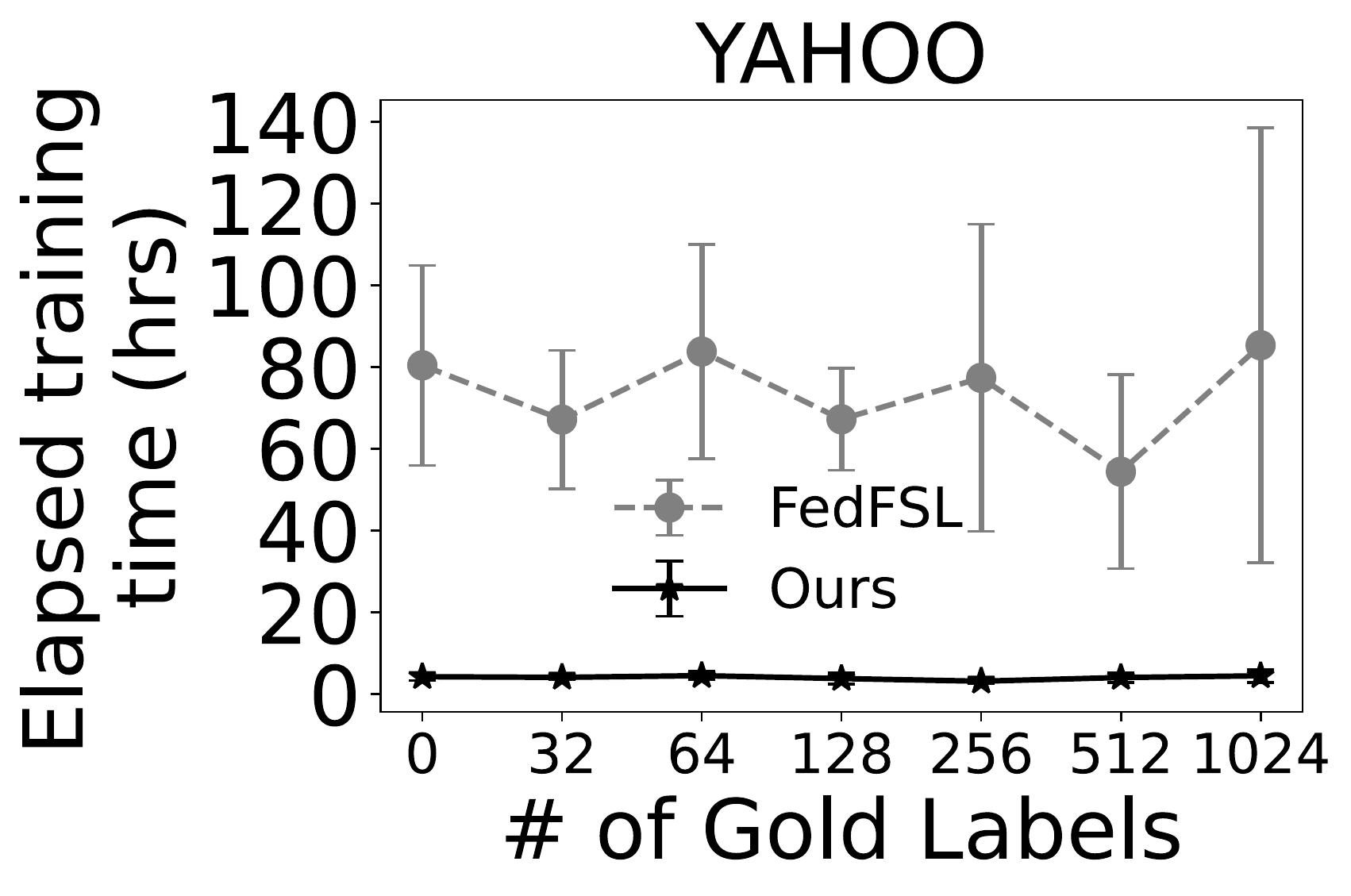}  \vspace{-0.5cm}
        \end{minipage}
        \vspace{-5pt}
        \subcaption{Training time}\label{fig:eval-overall-datapoints-cost}
    \end{minipage}
    \vspace{-10pt}
    \caption{The training accuracy (a) and training time (b) with different number of gold labels.
    }
    \label{fig:eval-overall-datapoints}
    \vspace{-10pt}
\end{figure*}

\textbf{\sys{} outperforms baselines in various network environments.}
Figure \ref*{fig:eval-overall-bandwidth} reports the performance of \sys and baselines under various network environments from 0.1MB/s to 10MB/s, which cover the typical WiFi and cellular bandwidth.
Our key observation is that \sys consistently outperforms other baselines under different network settings, with improvement more significant at lower bandwidths.
For instance, with a bandwidth of 10MB/s, \sys is $19.0\times$ and $33.6\times$ faster than \texttt{FedFSL} on \texttt{AGNEWS} and \texttt{MNLI}, respectively.
When the bandwidth drops to 0.1MB/s, the improvement reaches 224.3$\times$ and 661.7$\times$, respectively.
This is due to \sys significantly reducing network transmission by tuning very few parameters ($\thicksim$0.1\% of all.)

Our results show that \sys also outperforms \texttt{FedFSL-BIAS} by 5.6$\times$--7.4$\times$.
Those benefits arise from \sys{}'s greater efficiency in both computation and communication, which is also verified by the end-to-end performance.


\begin{figure}[t]
	\centering
        \includegraphics[width=0.5\textwidth]{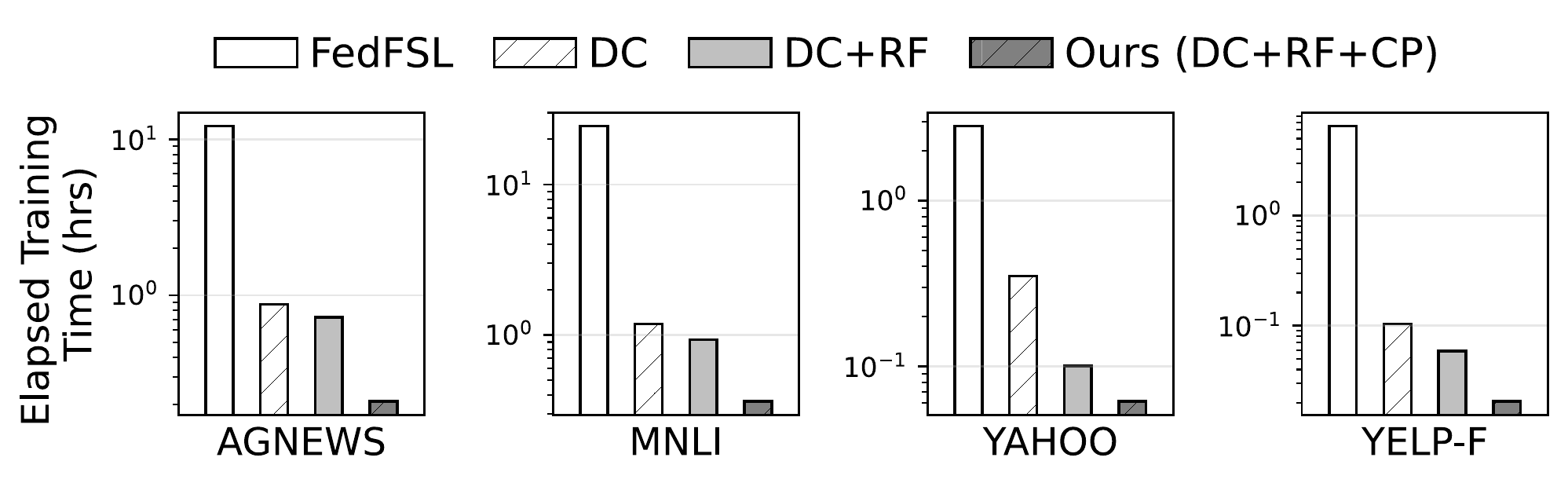}

	 \vspace{-15pt}
	\caption{
		Model convergence delays with and without \sys{}'s key designs, showing their significance. DC: training depth/capacity co-planning; RF: representative filtering; CP: curriculum pacing. 
        }
	\vspace{-10pt}
	\label{fig:eval-ablation-all}
\end{figure}

\subsection{Impacts of Initial Gold Labels} \label{sec:eval-datapoints}
We vary the initial data labels and compare the performance of \sys to baselines on two datasets: \texttt{AGNEWS} and \texttt{YAHOO}.
As shown in Figure~\ref{fig:eval-overall-datapoints-acc}, \sys performs on par with or slightly higher than \texttt{FedFSL} in terms of relative accuracy from 0--1024 initial data labels, which is up to 64.1\% higher than \texttt{FedCLS}.
In some cases, \sys achieves satisfactory zero-shot performance, e.g., 95.2\% relative accuracy on \texttt{AGNEWS} while \texttt{FedCSL} only reaches 31.1\%.
This observation paves the way for future research on zero-shot learning in mobile NLP.
Furthermore, \sys also significantly reduces the end-to-end convergence time under various initial data labels.
For a fair comparison, we only compare \texttt{FedFSL} and \sys that perform alike.
As shown in Figure~\ref{fig:eval-overall-datapoints-cost}, to reach the same accuracy, \sys reduces the elapsed training time by up to 18.3$\times$ and 17.1$\times$ on \texttt{AGNEWS} and \texttt{YAHOO}, respectively.

\subsection{Significance of Key Designs}
\label{sec:eval-ablation}

We perform an ablation study to understand the contribution of each key technique of \sys presented in $\S$\ref{sec:design}.
As shown in Figure~\ref{fig:eval-ablation-all}, we find each of them significantly contributes to the results: 
(1) The co-planning of training depth and capacity reduces the convergence time by 8.0$\times$--62.3$\times$ on different datasets.
The significant improvement comes from that most of bottom layers (up to 66.7\% on \texttt{AGNEWS}) are skipped, which reduces the training latency linearly.
Some middle layers (up to 33.3\%) are tuned with reduced capacity, and thus reduces the network traffic.
(2) With the model optimized for training, we observe the pseudo labeling accounts for more than 70\% of the total computation cost.
The representative diversity mechanism filters out up to 95\% of the data, further reducing the training time by 1.2$\times$--3.5$\times$.
(3) Curriculum pacing further reduces the training time by 1.6$\times$--3.5$\times$ by selecting a (sub-)optimal pacing configuration.

\subsection{Client Resource Cost}
\label{sec:eval-cost}
\begin{figure}[t]
	\centering
	\includegraphics[width=0.48\textwidth]{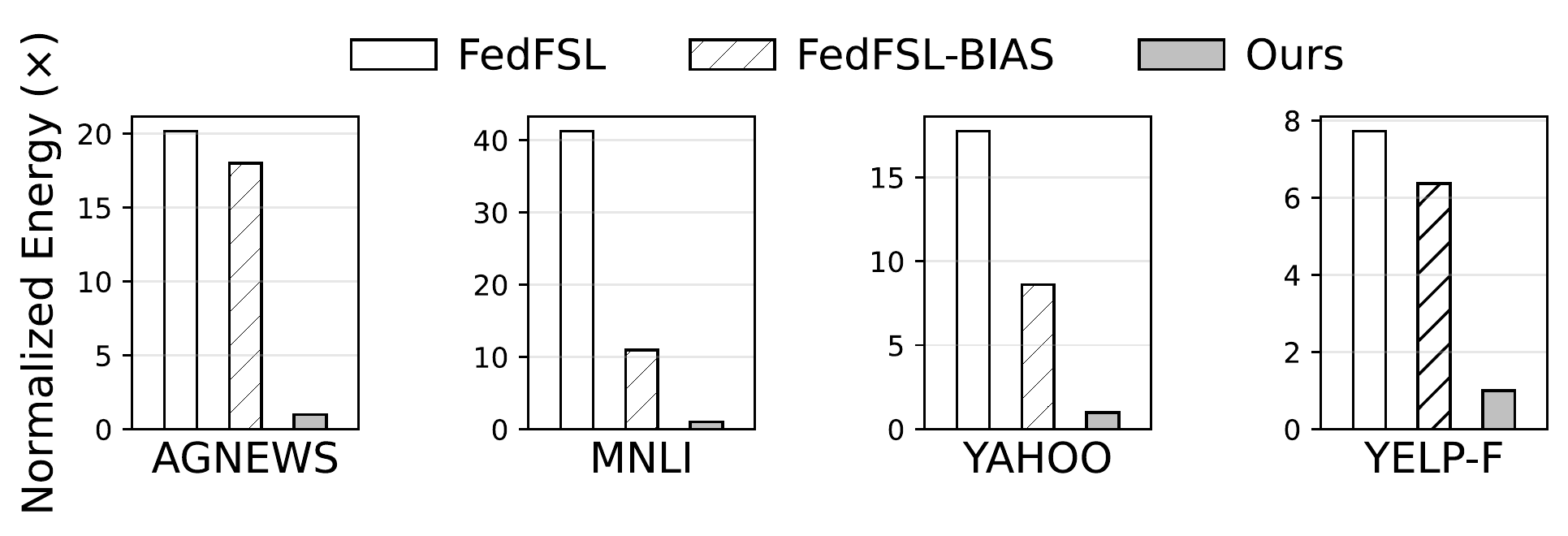}
	\vspace{-25pt}
	\caption{The total energy consumption of all clients, normalized to that of \texttt{FedCLS}.
    }
	\vspace{-10pt}
    \label{fig:eval-cost-energy}
\end{figure}

\paragraph{Energy consumption}
Figure~\ref{fig:eval-cost-energy} illustrates the average energy consumed during mobile NLP training tasks on each device.
It shows that \sys saves the energy consumption remarkably, e.g., 7.7$\times$--41.2$\times$ reduction compared to \texttt{FedFSL} and 6.4$\times$--18.0$\times$ reduction compared to \texttt{FedFSL-BIAS}.
This improvement comes from the reduced network transmission time, the on-device training/labeling computations, and the cherry-picked orchestrating pace.

\begin{figure}[t]
	\centering
	\includegraphics[width=0.48\textwidth]{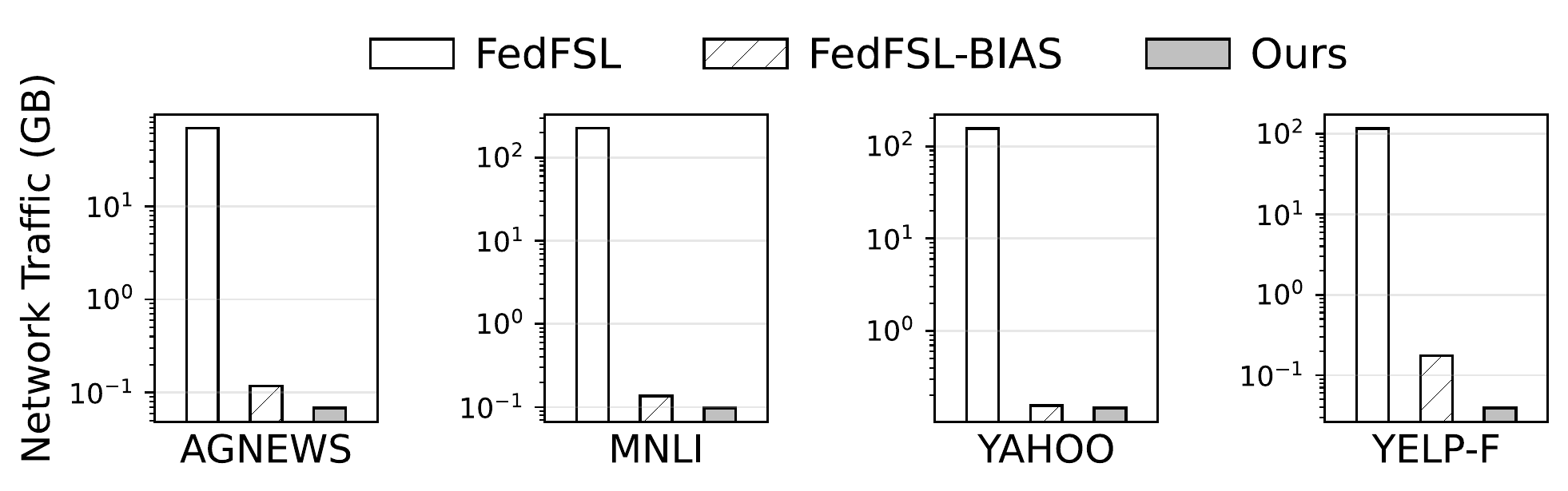}
	\vspace{-25pt}
	\caption{The total network traffic of all clients.
 }
    \label{fig:eval-cost network}
\end{figure}

\paragraph{Network traffic}
Figure \ref{fig:eval-cost network} reports the total network traffic incurred during fine-tuning to reach the convergence accuracy of \texttt{FedFSL}. 
It shows that \sys saves 1841.7$\times$ on average and up to $3000.0\times$ (reducing from 224.6 GB to 0.04 GB) network traffic compared to \texttt{FedFSL} on four datasets.
Please note that reducing the network traffic not only speeds up the convergence, 
but also mitigates the overhead on clients and the monetary cost to FL developers.
The cost is billed by the amount of data transmitted on public cloud platforms such as AWS~\cite{awsbill2022}, which charges \$0.01/GB.

\begin{figure}[t]
	\centering
	\includegraphics[width=0.35\textwidth]{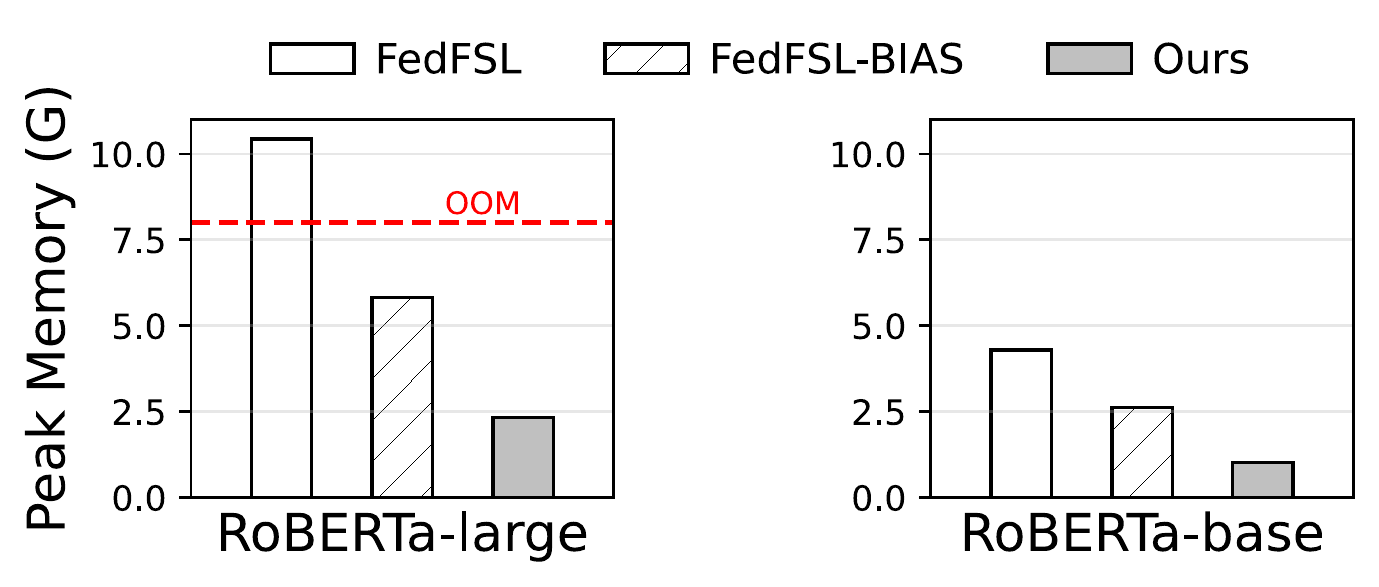}
	\vspace{-15pt}
	\caption{Memory footprint of on-device training.}
    \label{fig:eval-cost-memory}
	\vspace{-10pt}
\end{figure}

\paragraph{Memory footprint}
As shown in Figure~\ref{fig:eval-cost-memory}, our training depth and capacity co-planning mechanism can reduce the memory footprint by 4.3--4.5 times, which is crucial for practical deployment on mobile devices.
For example, \texttt{FedFSL} requires 10.4 GB memory\footnote{Tested on a central server.} to train RoBERTa-large, which is 2.4$\times$ higher than training RoBERTa-base.
This excessive memory requirement would lead to out-of-memory and training failure on mobile devices which typically have only 8GB RAM.
FedFSL-BIAS reduces the memory usage of training RoBERTa-large to 5.8 GB, which is still too large for mobile devices.
Because it only bypasses the memory bottleneck of the weight update, but not the intermediate activations which is the main memory bottleneck~\cite{cai2020tinytl}.
In comparison, \sys only requires 2.3 GB memory due to the shallow training depth and greatly saved intermediate activations. 

\paragraph{Remark}
Training can be done when no user interactions are present, e.g. when phone is idle/charged overnight which is nearly a “clean” environment without other co-running applications to share memory. 
Moreover, memory inefficiency can be compensated with acceptable training overhead through advanced memory optimizations such as batch splitting and model weight caching~\cite{wang2022melon}.
During the end-to-end convergence, which typically takes between 0.1 to 0.8 hours, each device typically engages in a few rounds of training, with each round lasting only a few tens of seconds. 
As a result, \sys shall not compromise user experience.

    \section{Related Work}\label{sec:related}

\paragraph{Few-shot learning (FSL) and FedFSL}
FSL has been one of the hottest topics in machine learning research, as it is considered more akin to how human intelligence works~\cite{wang2020generalizing, sung2018learning, ravi2016optimization,snell2017prototypical, garcia2017few}.
\sys identifies two complementary algorithmic blocks, i.e., pseudo labeling~\cite{zhang2021flexmatch, cascante2021curriculum,lee2013pseudo, arazo2020pseudo} and prompt learning~\cite{liu2021pre}, and demonstrates satisfactory accuracy under federated context.
Prior work~\cite{https://doi.org/10.48550/arxiv.2204.03649} introduced iterative pseudo labeling into prompt learning for centralized training of vision-language model, but it employed a fixed iterative pace that could result in poor federated performance, as we have demonstrated.
Our harsh but practical assumptions of highly scare and skewed data labels invalidates existing FedFSL methods~\cite{zhao2020semi, feng2022semi, diao2021semifl, jeong2020federated,fan2021federated, yu2021fedhar}, which assume at least 4\% of the data to be uniformly labeled across clients.
As far as we know, \sys achieves the state-of-the-art performance in the FedFSL scenario.
At system aspect, FSL is generally considered lightweight due to fewer rounds of training~\cite{fan2021federated, zhang2020revisiting, guha2019one, chen2019closer}.
However, we show that the cost could be substantial in FL due to the use of the pseudo labeling (requiring on-device inference) and prompt learning (requiring large NLP models).
We are the first to tackle these system challenges and enable practical FedFSL.

\paragraph{Large language model (LLMs)}
Few-shot prompt tuning on LLMs such as GPT-3~\cite{brown2020language} can rival the performance of fully-supervised medium-size models like RoBERTa~\cite{liu2019roberta}.
However, these models are too large (e.g., 175B+ parameters for GPT-3) to be deployed on devices after training, and deploying them on the cloud leads to privacy concerns~\cite{sweeney2000simple, wikipedia} and network delays~\cite{xu2021cloud, haq2017measuring}.
We pioneer practical federated prompt tuning, allowing resource-constrained devices to achieve comparable few-shot performance while preserving privacy and supporting offline inference.

\paragraph{FedNLP} aims to achieve both high accuracy and privacy preservation in NLP model fine-tuning.
Recently, there are a few literature investigating its implications, but mostly at the algorithm aspect.
\cite{lin-etal-2022-fednlp} builds a benchmark for popular FedNLP tasks and datasets in a standard FL workflow.
\cite{basu2021benchmarking} enhances the privacy of FedNLP by orchestrating with differential privacy.
SEFL~\cite{deng2022secure} eliminates the need for the trusted entities and is resilient to client dropouts in FedNLP tasks.
Those work are orthogonal to \sys.
\cite{cai2022autofednlp} is the only work that we are aware of that tackles with huge system cost of FedNLP.
It proposes a FedNLP framework based on lightweight, automatically configured adapters at runtime.
However, the adapter cannot be applied in few-shot NLP scenarios according to our experiments.

\paragraph{FL system optimizations}
The huge resource cost of cross-device FL has been well recognized by the research community.
In respond, lots of efforts have been invested, including communication efficiency optimizations~\cite{bonawitz2019towards1,yang2021characterizing}, model compression/quantization~\cite{wu2018error, bernstein2018signsgd}, client/data sampling~\cite{li2021hermes,lipyramidfl, nishio2019client, xu2020client, wang2021device, lai2020oort, zhao2021quality, li2021sample}, and on-device training speedup~\cite{xu2022mandheling,wang2022melon}.
Instead, \sys addresses unique challenges raised by the few-shot scenarios: pacing between pseudo labeling and training; filtering redundant unlabeled data for pseudo labeling.
The design of \sys is mostly compatible with most optimizations above as its FL training is loosely coupled with the pseudo labeling.

\paragraph{Attacks in FL}
It is well known that FL cannot fully guarantee privacy preservation, e.g., extraction attacks~\cite{deng2021tag, zhu2019deep, balunovic2022lamp}.
However, dropout, a common training technique used in ML, is proven to be very effective to defend against those attacks~\cite{deng2021tag}.
Moreover, \cite{zhu2019deep} demonstrates that most attacks suffer a significant decrease in success ratio when training batch sizes are set greater than 1.
Apart from that, most data extraction attacks tend to be extremely resource-intensive~\cite{deng2021tag, zhu2019deep, balunovic2022lamp}. 
Though larger models leak more information than the smaller ones, it incur larger inversion cost either (e.g., about 1672.52s for reconstructing one sentence~\cite{zhu2019deep}). 
\sys avoids revealing training data and raises the barrier for attackers (i.e. it requires much higher attack capability and much longer time).
Furthermore, integrating various privacy-preserving techniques, such as differential privacy~\cite{basu2021benchmarking} and secure aggregation~\cite{bonawitz2017practical}, can further enhance the security of FL.
\sys is parameter-efficient and thereby shall be easy to integrate with them.
    \section{Conclusions}\label{sec:conclusions}

\sys is a FedFSL framework that enables practical few-shot NLP fine-tuning on federated mobile devices.
At algorithm aspect, it incorporates pseudo labeling and prompt learning to achieve usable accuracy with only tens of data labels.
At system aspect, it proposes three novel techniques, i.e., by pacing training and labeling, early filtering unlabeled data, and reducing the tuning depth/capacity, to address the unique challenge of huge resource cost raised by its algorithmic foundation.
On extensive experiments, \sys shows superior system performance over existing approaches.

\section*{Acknowledgments}
This research was supported by National Key Research and Development Program of China \#2020YFB1805500, NSFC \#62032003, \#61921003, \#62102045, Beijing Nova Program \#Z211100002121118, Young Elite Scientists Sponsorship Program by CAST \#2021QNRC001, and CCF-Alibaba Innovative Research (AIR). 
Dongqi Cai was supported by BUPT Excellent Ph.D. Students Foundation \#CX2023124.
The authors thank the anonymous reviewers and the shepherd for their insightful feedbacks.
    
    \balance
    \bibliographystyle{plain}
    \bibliography{bib/ref-mwx,bib/nlp,bib/ref-cdq,bib/autofednlp}

\end{document}